\documentclass[letterpaper, 10 pt, conference]{ieeeconf}  

\IEEEoverridecommandlockouts                              

\usepackage{epsfig} 

\usepackage{graphicx}
\usepackage{amsmath}
\usepackage{amssymb}
\usepackage{gensymb}
\usepackage[font={small}]{caption}
\usepackage{subcaption}
\usepackage{hhline}
\usepackage{afterpage}
\usepackage{float}
\usepackage{wrapfig}
\usepackage{placeins}
\usepackage{booktabs} 
\usepackage{algorithm}
\usepackage{algpseudocode}

\usepackage{hyperref}

\graphicspath{{Figs/}{Figs/hardest/}{Figs/histograms/}{Figs/overview/}{Figs/confusion_matrix/}}

\title{\LARGE \bf
Modelling Uncertainty in Deep Learning for Camera Relocalization
}

\author{Alex Kendall and Roberto Cipolla
\thanks{The authors are with the Department of Engineering, University of Cambridge, UK. Contact: {\tt\small \href{mailto:agk34@cam.ac.uk}{agk34@cam.ac.uk}} \newline The web demo, code and dataset can be viewed on the project webpage {\tt\small \href{http://mi.eng.cam.ac.uk/projects/relocalisation/}{mi.eng.cam.ac.uk/projects/relocalisation/}}}%
}

\begin{document}

\maketitle
\thispagestyle{empty}
\pagestyle{empty}

\begin{abstract}
We present a robust and real-time monocular six degree of freedom visual relocalization system. We use a Bayesian convolutional neural network to regress the 6-DOF camera pose from a single RGB image. It is trained in an end-to-end manner with no need of additional engineering or graph optimisation. The algorithm can operate indoors and outdoors in real time, taking under 6ms to compute. It obtains approximately 2m and 6\degree accuracy for very large scale outdoor scenes and 0.5m and 10\degree accuracy indoors. Using a Bayesian convolutional neural network implementation we obtain an estimate of the model's relocalization uncertainty and improve state of the art localization accuracy on a large scale outdoor dataset. We leverage the uncertainty measure to estimate metric relocalization error and to detect the presence or absence of the scene in the input image. We show that the model's uncertainty is caused by images being dissimilar to the training dataset in either pose or appearance.
\end{abstract}


\section{Motivation}

Modern Simultaneous Localization and Mapping (SLAM) systems perform well for a number of applications such as augmented reality and domestic robotics \cite{klein2007parallel}. However, we are yet to see their wide spread use in the wild because of their inability to cope with large viewpoint or appearance changes. The point landmarks used by visual SLAM, such as SIFT \cite{lowe2004distinctive} or ORB \cite{rublee2011orb}, are not able to create a representation which is sufficiently robust to these challenging scenarios. Dense SLAM systems \cite{newcombe2011dtam,engel2014lsd} attempt to estimate camera motion directly from the pixels, but are also fragile in these situations. Their assumption of small changes between frames breaks with large viewpoint or appearance changes. In addition to these short comings, both these metric SLAM frameworks require a good initial pose estimate to be able to track the camera's pose continuously and are ineffective at relocalization.

A complimentary system, appearance based SLAM, can relocalize to a coarse pose estimate, but is constrained to classifying the scene among a limited number of discrete locations. This has been achieved using SIFT \cite{lowe2004distinctive} features with FAB-MAP \cite{cummins2008fab} or using convolutional neural network features \cite{sunderhauf2015place}.

In \cite{kendall2015convolutional}, we introduced a new framework for localization, PoseNet, which overcomes many limitations of these current systems. It removes the need for separate mechanisms for appearance based relocalization and metric pose estimation. Furthermore it does not need to store key frames, or establish frame to frame correspondence. It does this by mapping monocular images to a high dimensional space which is linear in pose and robust to annoyance variables. We can then regress the full 6-DOF camera pose from this representation. This allows us to regress camera pose from the image directly, without the need of tracking or landmark matching.

\begin{figure}[t]
\begin{center}
	\includegraphics[width=\linewidth]{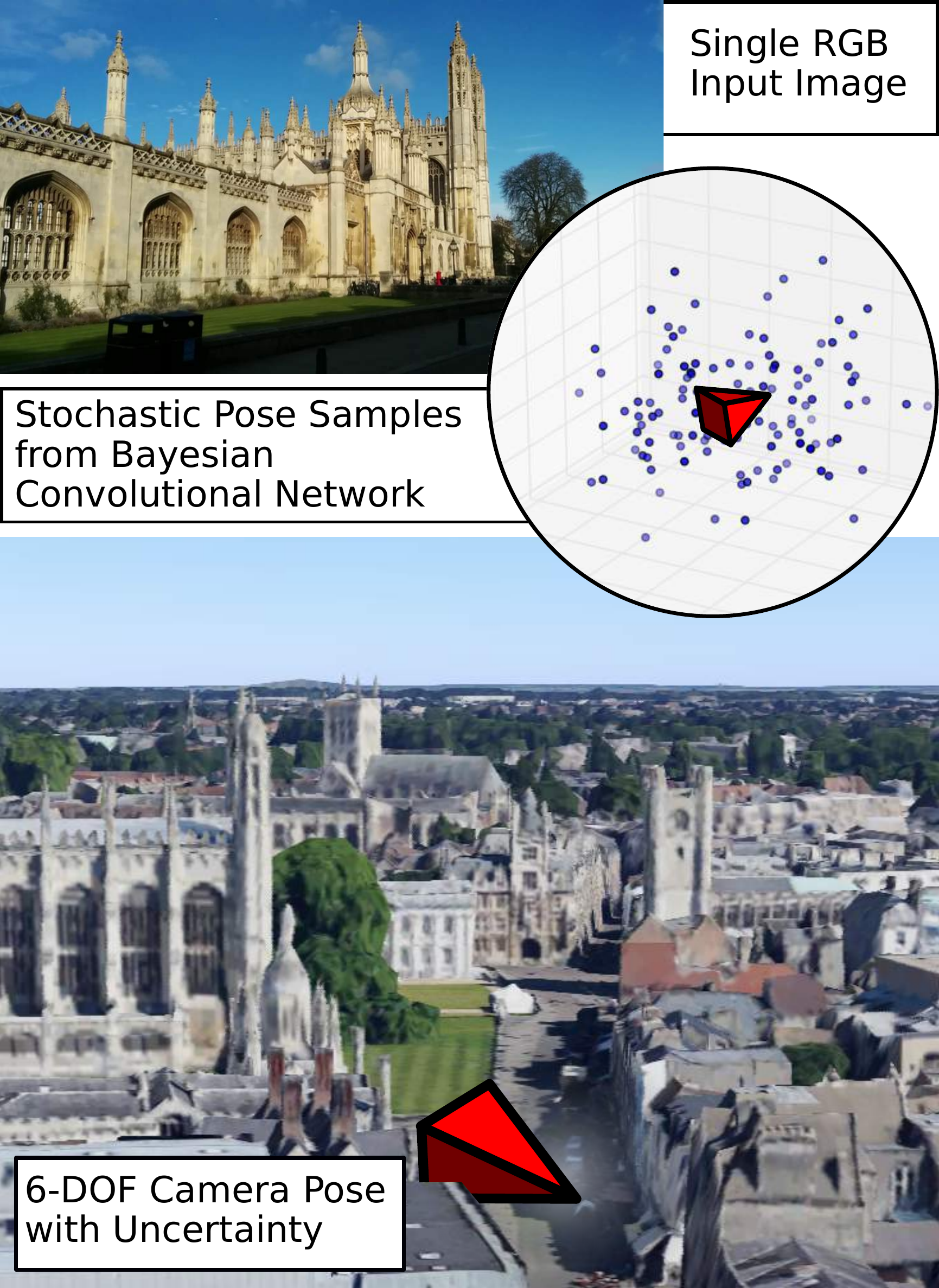}
\end{center}
   \caption{\textbf{Bayesian convolutional network 6-DOF pose regressor.} Our system takes a single RGB image as input and outputs the 6-DOF camera pose with a measure of model uncertainty. The system operates in under $6$ms per image on a GPU, requiring $<50$MB memory and can relocalize within approximately $2$m and $6\degree$ for large outdoor scenes spanning $50, 000$m$^2$. For an online demonstration, please see {\tt\small\href{http://mi.eng.cam.ac.uk/projects/relocalisation/}{mi.eng.cam.ac.uk/projects/relocalisation/}}}
\label{fig:overview}
\end{figure}

The main contribution of this paper is extending this framework to a Bayesian model which is able to determine the uncertainty of localization. Our Bayesian convolutional neural network requires no additional memory, and can relocalize in under 6ms per frame on a GPU. By leveraging this probabilistic approach, we achieve an improvement on PoseNet's performance on \textit{Cambridge Landmarks}, a very large urban relocalization dataset, and \textit{7 Scenes}, a challenging indoor relocalization dataset. Furthermore, our approach qualitatively improves the system by producing a measure of model uncertainty. We leverage this uncertainty value to estimate:
\begin{itemize}
\item metric relocalization error for both position and orientation,
\item the confidence of modelling the data (detect if the scene is actually present in the input image).
\end{itemize}

Understanding model uncertainty is an important feature for a localization system. A non-Bayesian system which outputs point estimates does not interpret if the model is making sensible predictions or just guessing at random. By measuring uncertainty we can understand with what confidence we can trust the prediction.

Secondly, it is easy to imagine visually similar landmarks and we need to be able to understand with what confidence can we trust the result. For example, there are many examples of buildings with visually ambiguous structures, such as window, which are tessellated along a wall.

\section{Related Work}

Visual SLAM \cite{klein2007parallel,newcombe2011dtam,engel2014lsd,mur2015orb}, and relocalization research \cite{wang2006coarse,li2012worldwide}, has largely focused on registering viewpoints using point-based landmark features, such as SIFT \cite{lowe2004distinctive} or ORB \cite{rublee2011orb}. The problem with using local point-based features, is that they are not robust to many real-life scenarios. Tracking loss typically fails with rapid motion, large changes in viewpoint, or significant appearance changes.

Convolutional neural networks have recently been shown to be extremely powerful for generating image representations for classification \cite{krizhevsky2012imagenet,szegedy2014going}. Their hierarchical structure has been shown to build representations from an object level. These higher level contours and shape cues \cite{zeiler2014visualizing} have been shown to be increasingly invariant to object instantiation parameters. However their use for regression has been predominantly limited to object detection \cite{szegedy2014going} and human pose regression \cite{toshev2014deeppose}. Our approach \cite{kendall2015convolutional} leverages convolutional network's advanced high level representations to regress the full 3D camera pose. We show in \cite{kendall2015convolutional} that using these high-level features allows us to localize with unknown camera intrinsics, different weather, silhouette lighting and other challenging scenarios.

A learning approach to relocalization has also been proposed previously using random forests to regress Scene Coordinate labels \cite{shotton2013scene}. It was extended with a probabilistic approach in \cite{valentin2015exploiting}. However this algorithm has not been demonstrated beyond a scale of a few $m^3$ and requires RGB-D images to generate the scene coordinate label, in practice constraining its use to indoor scenes.

Although many of the modern SLAM algorithms do not consider localization uncertainty \cite{klein2007parallel,engel2014lsd,li2012worldwide}, previous probabilistic algorithms have been proposed. Bayesian approaches include extended Kalman filters and particle filter approaches such as FastSLAM \cite{thrun2005probabilistic}. However these approaches estimate uncertainty from sensor noise models, not the uncertainty of the model to represent the data. Our proposed framework does not assume any input noise but measures the model uncertainty for localization.

Neural networks which consider uncertainty are known as Bayesian neural networks \cite{denker1991transforming,mackay1992practical}. They offer a probabilistic interpretation of deep learning models by inferring distributions over the networks’ weights. They are often very computationally expensive, increasing the number of model parameters without increasing model capacity significantly. Performing inference in Bayesian neural networks is a difficult task, and approximations to the model posterior are often used, such as variational inference \cite{graves2011practical}.

On the other hand, the already significant parameterization of convolutional network architectures leaves them particularly susceptible to over-fitting without large amounts of training data. A technique known as \textit{dropout} is commonly used as a regularizer in convolutional neural networks to prevent overfitting and coadaption of features \cite{srivastava2014dropout}. During training with stochastic gradient descent, \textit{dropout} randomly removes connections within a network. By doing this it samples from a number of thinned networks with reduced width. At test time, standard dropout approximates the effect of averaging the predictions of all these thinnned networks by using the weights of the unthinned network. This can be thought of as sampling from a distribution over models.

Recent work \cite{Gal2015DropoutB} has extended dropout to approximate Bayesian inference over the network's weights. Gal and Ghahramani \cite{Gal2015Bayesian} show that dropout can be used at test time to impose a Bernoulli distribution over the convolutional net filter's weights, without requiring any additional model parameters. This is achieved by sampling the network with randomly dropped out connections at test time. We can consider these as Monte Carlo samples which sample from the posterior distribution of models.

This is significantly different to the `probabilities' obtained from a softmax classifier in classification networks. The softmax function provides relative probabilities between the class labels, but not an overall measure of the model's uncertainty.

\section{Model for Deep Regression of Camera Pose}

For clarity, we briefly review PoseNet which we proposed in \cite{kendall2015convolutional} to learn camera pose with deep regression. We use a modified GoogLeNet architecture \cite{szegedy2014going} which is a 23 layer deep convolutional neural network. We replace the classifiers with pose regressors, representing 3D pose with position $\mathbf{x}$ and orientation $\mathbf{q}$, expressed as a quaternion. We train the network to regress this 7 dimensional vector with the following Euclidean loss function:
\begin{equation}
loss(I) = \left\lVert\mathbf{\hat{x}} - \mathbf{x}\right\rVert_2 + \beta \left\lVert \mathbf{\hat{q}}-\frac{\mathbf{q}}{\left\lVert\mathbf{q}\right\rVert}\right\rVert_2
\label{eqn:loss}
\end{equation}
The parameter $\beta$ is tuned to optimally learn both position and orientation simultaneously. We train the model end-to-end with stochastic gradient descent. We show in \cite{kendall2015convolutional} that with transfer learning we can train the system using only a few dozen training examples.

\section{Modelling Localization Uncertainty}

\begin{figure*}[t]
\begin{center}
	\begin{subfigure}{0.25\linewidth}
		\begin{center}
        \includegraphics[width=\linewidth]{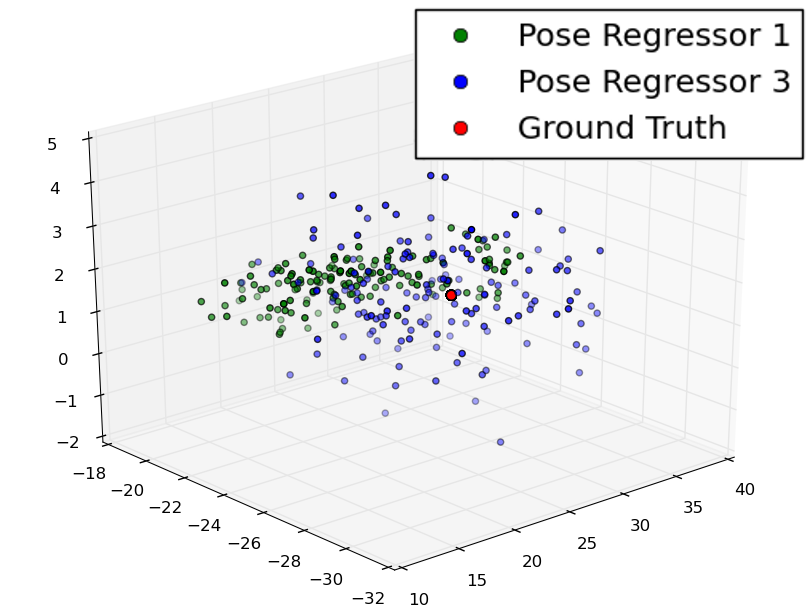}
        \includegraphics[width=0.5\linewidth]{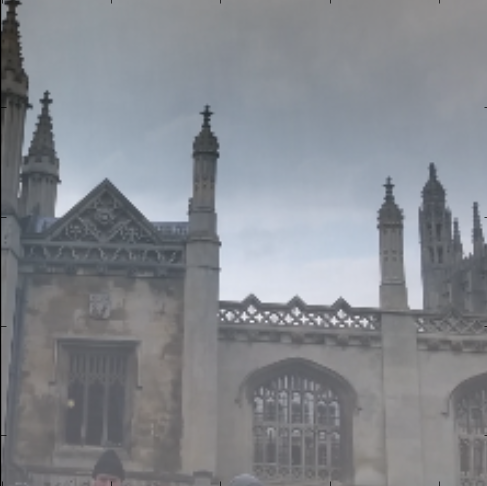}
		\end{center}
        \caption{King's College}
    \end{subfigure}
    	\begin{subfigure}{0.25\linewidth}
		\begin{center}
        \includegraphics[width=\linewidth]{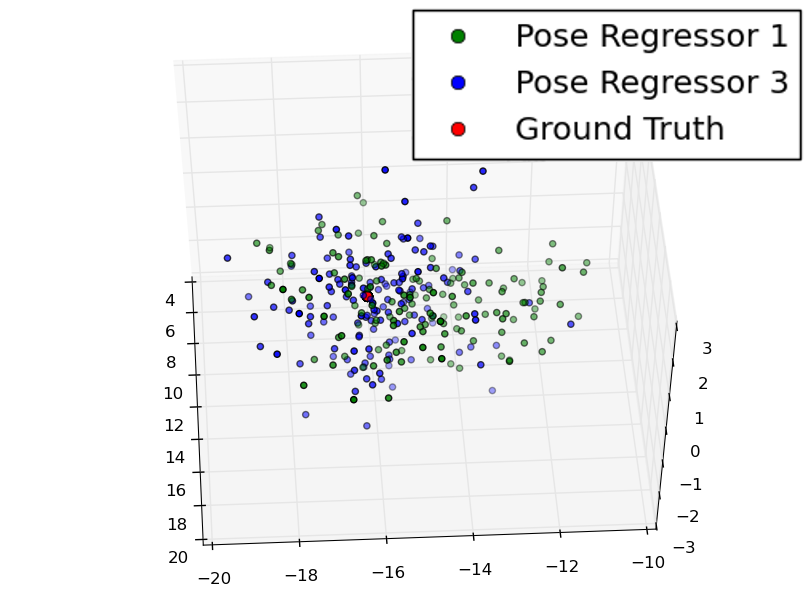}
        \includegraphics[width=0.5\linewidth]{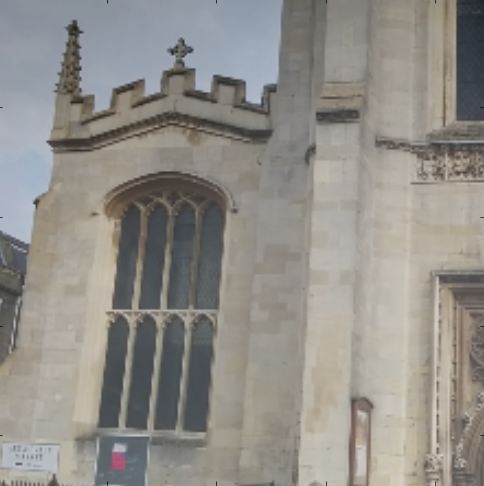}
		\end{center}
        \caption{St Mary's Church}
    \end{subfigure}
	\begin{subfigure}{0.25\linewidth}
		\begin{center}
        \includegraphics[width=\linewidth]{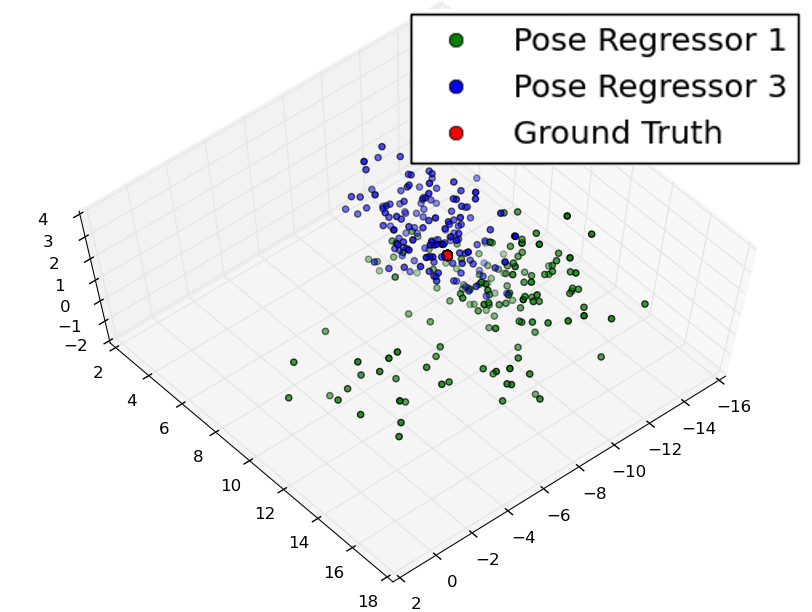}
        \includegraphics[width=0.5\linewidth]{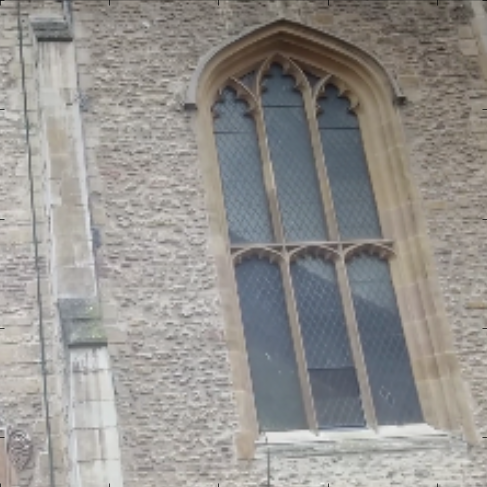}
		\end{center}
        \caption{St Mary's Church.}
    \end{subfigure}
\end{center}
   \caption{3D scatter plots of \textbf{Monte Carlo pose samples from the Bayesian convolutional neural network} (top row) from an input image (bottom row)  from the posterior distribution. We show typical examples from two scenes (a,b) and a visually ambiguous example (c). In green are the results from the first auxiliary pose regressor and in blue are samples from the final pose regressor. It shows that the auxiliary pose predictions (from the shallower sub-net) are typically multimodal however results from the final regressor are unimodal.}
\label{fig:pose_samples_3d}
\end{figure*}

Gal and Ghahramani \cite{Gal2015Bayesian} have shown a theoretical relationship between dropout \cite{srivastava2014dropout} and approximate inference in a Gaussian process. We can consider sampling with dropout as a way of getting samples from the posterior distribution of models. They link this approximate model to variational inference in Bayesian convolutional neural networks with Bernoulli distributions. We leverage this method to obtain probabilistic inference of our pose regression model, forming a Bayesian PoseNet.

Firstly, we briefly summarise the method to obtain a Bayesian convolutional neural network introduced by \cite{Gal2015Bayesian}. We are interested in finding the posterior distribution over the convolutional weights, $\mathbf{W}$, given our observed training data $\mathbf{X}$ and labels $\mathbf{Y}$.
\begin{equation}
p(\mathbf{W}~|~\mathbf{X},\mathbf{Y})
\end{equation}
In general, this posterior distribution is not tractable, we must use some learning method to approximate the distribution of these weights \cite{denker1991transforming}. We use variational inference to approximate it \cite{graves2011practical}. This technique allows us to learn the distribution over the network's weights, $q(\mathbf{W})$, by minimising the Kullback-Leibler (KL) divergence between this approximating distribution and full posterior.

\begin{equation}
\text{KL}(q(\mathbf{W})~||~p(\mathbf{W}~|~\mathbf{X},\mathbf{Y}))
\end{equation}
Where the approximating variational distribution $q(\mathbf{W_i})$ for every layer $i$ is defined as:
\begin{equation}
\begin{split}
\mathbf{b}_{i,j} \sim \text{Bernoulli}(p_i) \text{ for } j = 1, ..., K_{i-1} \\
\mathbf{W_i} = \mathbf{M}_i \text{diag}(b_i)
\end{split}
\end{equation}
with $b_i$ vectors of Bernoulli distributed random variables and variational parameters $\mathbf{M_i}$ we obtain the approximate model of the Gaussian process in \cite{Gal2015Bayesian}. The dropout probabilities, $p_i$, could be optimised. However we leave them at the standard probability of dropping a connection as 50\%, i.e. $p_i=0.5$ \cite{srivastava2014dropout}.

\cite{Gal2015Bayesian} shows that minimising the Euclidean loss objective function in the pose regression network has the effect of minimising the Kullback-Leibler divergence term. Therefore training the network with stochastic gradient descent will encourage the model to learn a distribution of weights which explains the data well while preventing over-fitting.

As a result of this dropout interpretation of Bayesian convolutional neural networks, a dropout layer should be added after every convolutional layer in the network. However in practice this is not the case as is explained in section \ref{ch:arch}. Using standard libraries such as \cite{jia2014caffe} we can obtain an efficient implementation of a Bernoulli Bayesian convolutional neural network. At test time we perform inference by averaging stochastic samples from the dropout network.

Therefore the final algorithm for the probabilistic pose net is as follows:

\begin{algorithm}
\caption{Probabilistic PoseNet}
\begin{algorithmic}[1]
\Require image, learned weights $\mathbf{W}$, number of samples
\For{sample $= 1$ \textbf{to} number of samples}
 \State set network's weights to learned values
 \For{each weight \textbf{in} network}
  \State with probability $p$ set neuron activation to zero
 \EndFor
 \State samples $\leftarrow$ evaluate network
\EndFor
\State compute pose as average of all samples
\State compute uncertainty as a function of the samples' variance
\Ensure 6-DOF pose, uncertainty
\end{algorithmic}
\end{algorithm}

We also explored the possibility of using dense sliding window evaluation of the convolutional pose regressor over the input image to obtain a distribution of poses. This was done by taking $224\times224$ crops at regular intervals from the $455\times256$ pixel input image. This is equivalent to the densely evaluated PoseNet introduced in section \cite{kendall2015convolutional}. The variance of these pose samples also correlates with localization error, however not as strongly as sampling from a Bernoulli distribution over the weights.

\subsection{Estimating Uncertainty}

We can evaluate the posterior pose distribution from the Bayesian convolutional network by integrating with Monte Carlo sampling. Figure \ref{fig:pose_samples_3d} shows a plot of Monte Carlo samples from the output of the posterior network in blue. Additionally, in green we show the output from the first auxiliary pose regressor from the GoogLeNet architecture (see Figure 3 of \cite{szegedy2014going}). This output regresses pose from the representation after the inception (sub-net) layer 3. This result is at a much shallower depth and provides an insight as to what the network learns with additional depth. A similar result can be observed for the quaternion samples for the rotational component of pose.

For the full network's output (blue) we obtain a distribution that appears to be drawn from both an isotropic and single-modal Gaussian. The network appears to be very certain about the specific pose. By sampling with dropout over the distribution of models we observe some isotropic scatter around a single pose estimate.

At a shallower depth, with the first auxiliary pose regressor (green), the results are multi-modal. This is especially true for visually ambiguous images such as (c) in Figure \ref{fig:pose_samples_3d}. The window in image (c) is repeated along the face of St Mary's Church. Using dropout to sample the distribution of models at this shallower depth produces distributions which have components drawn from multiple pose hypotheses. This suggests that this extra depth in the network is able to learn a representation that is sufficiently discriminative to distinguish visually similar landmarks.

Therefore, we fit a unimodal Gaussian to the samples from the network's final pose regressor. We treat the mean of these samples as the point estimate for pose. For an uncertainty measurement we take the trace of the unimodal Gaussian's covariance matrix. We have found the trace to be an effective scalar measure of uncertainty. The trace is a sum of the eigenvalues, which is rotationally invariant and represents the uncertainty that the Gaussian contains effectively. Figure \ref{fig:error_vs_uncertainty} empirically shows this uncertainty measure is strongly correlated with metric error in relocalization.

We also considered using the determinant, which is a measure of the Gaussian's volume. However the determinant takes the product of the eigenvalues which means that if some are large and others are small then the resulting value will be small. This was  the case as the resulting Gaussian often had a strong elongated component to it, as can be observed in Figure \ref{fig:pose_samples_3d}. We found that using the determinant resulted in a numerically poorer measure of uncertainty.

We tried other models which accounted for multi-modality in the data:
\begin{itemize}
\item taking the geometric median instead of the mean as a point prediction,
\item fitting a mixture of Gaussians model to the data using the Dirichlet process \cite{blei2006variational},
\item clustering the samples using k-means and taking the mean of the largest cluster.
\end{itemize}
However we found that all of these methods produced poorer localization uncertainty than fitting a single unimodal Gaussian to the data.

\subsection{Creating a Comparable Uncertainty Statistic}
\label{ch:uncertainty_dist}

\begin{figure}
\begin{center}
	\begin{subfigure}[b]{0.49\linewidth}
        \includegraphics[width=\linewidth]{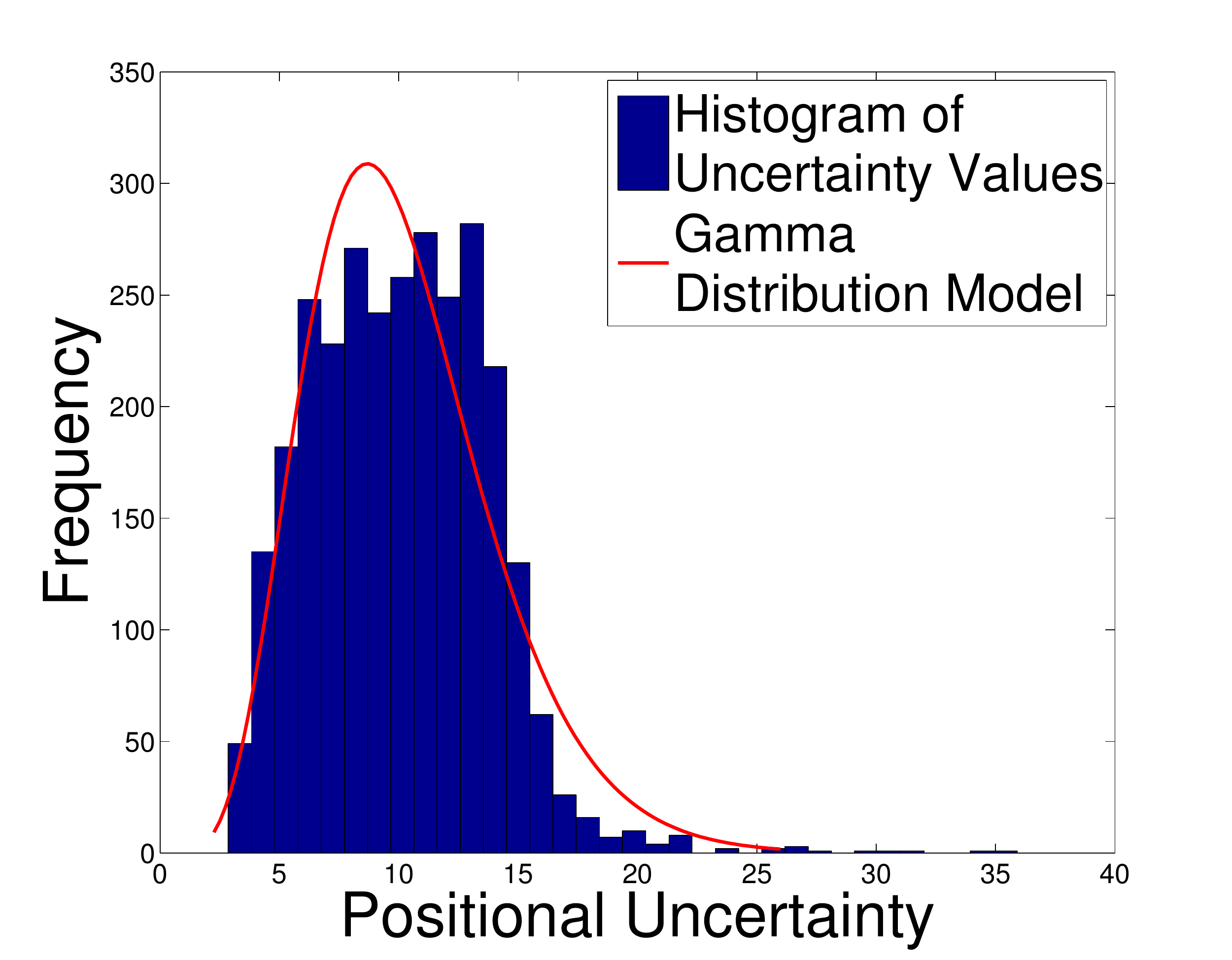}
        \caption{Translational uncertainty}
    \end{subfigure}
    	\begin{subfigure}[b]{0.49\linewidth}
        \includegraphics[width=\linewidth]{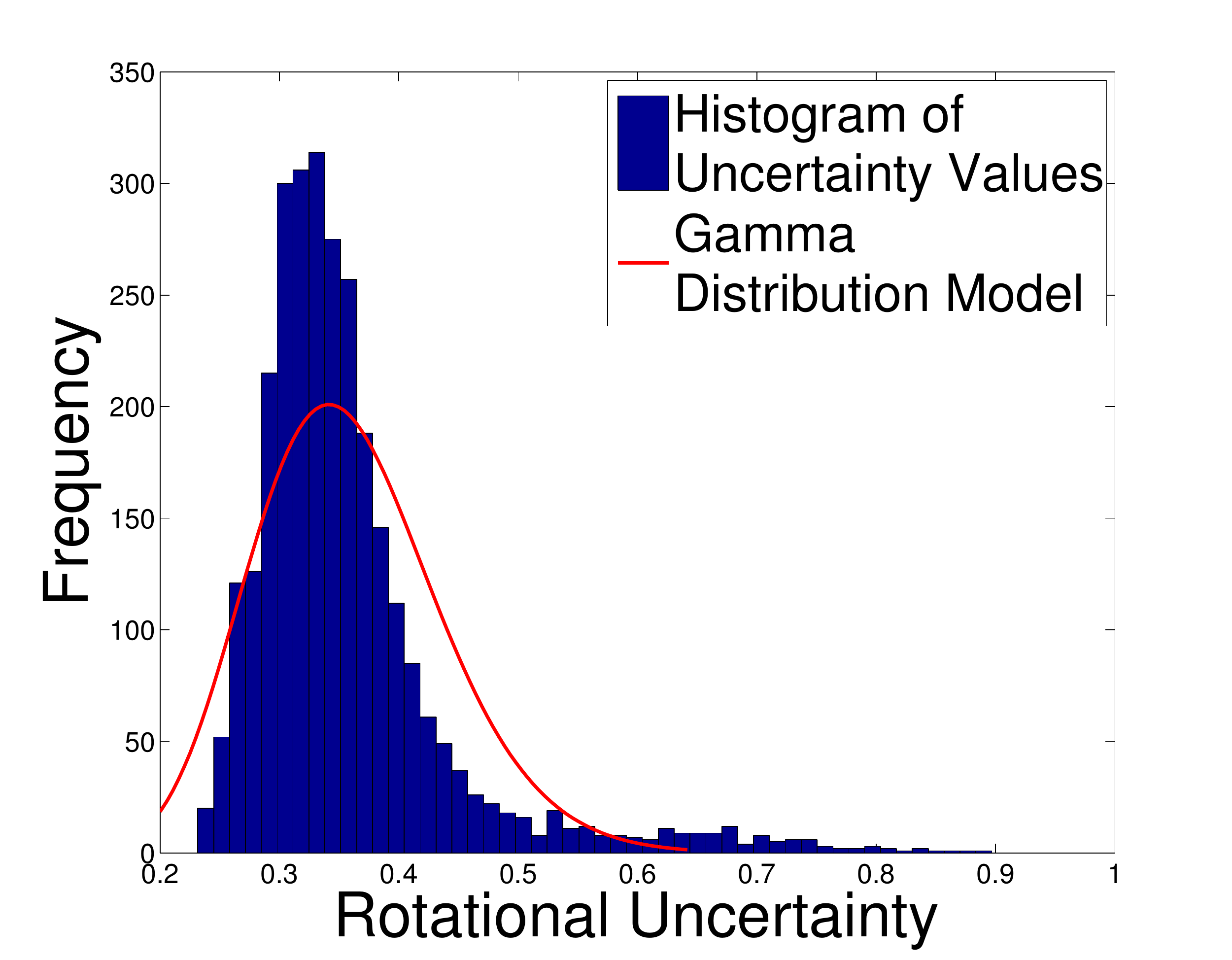}
        \caption{Rotational uncertainty}
    \end{subfigure}
\end{center}
   \caption{\textbf{Histograms of uncertainty values from the testing images in the Street scene.} In red we show the Gamma distribution used to model these populations. The Gamma distribution is a reasonable fit of the positively distributed, right skewed data.}
\label{fig:uncertainty_distribution}
\end{figure}

In order to compare the uncertainty values we obtained from a model, we propose the following method to create a normalized measure, or Z-score. This is an uncertainty value which is able to be directly compared between models.

To achieve this, firstly we evaluate the test dataset and record the predicted camera poses and associated uncertainties for each scene. Typical distribution of uncertainty results for the Street scene can be viewed in Figure \ref{fig:uncertainty_distribution}. Examining this distribution, we chose to model it with a Gamma distribution for three reasons; it requires only two parameters, the distribution is constrained to strictly positive values only and is right skewed.

Obtaining an estimate for the distribution of model uncertainty values for images from a scene's test set allows us to evaluate where a new image's uncertainty values sit compared to the population. We can now assign a percentile to both the translational and rotational uncertainty values by using the cumulative distribution function for the Gamma distribution. We treat this percentile as a Z-score and generate this from a separate distribution for both the translational and rotational uncertainties, as well as separately for each scene.

\begin{figure}[t]
\begin{center}
\makebox[\linewidth][c]{
	\begin{subfigure}[b]{0.33\linewidth}
        \includegraphics[width=\linewidth]{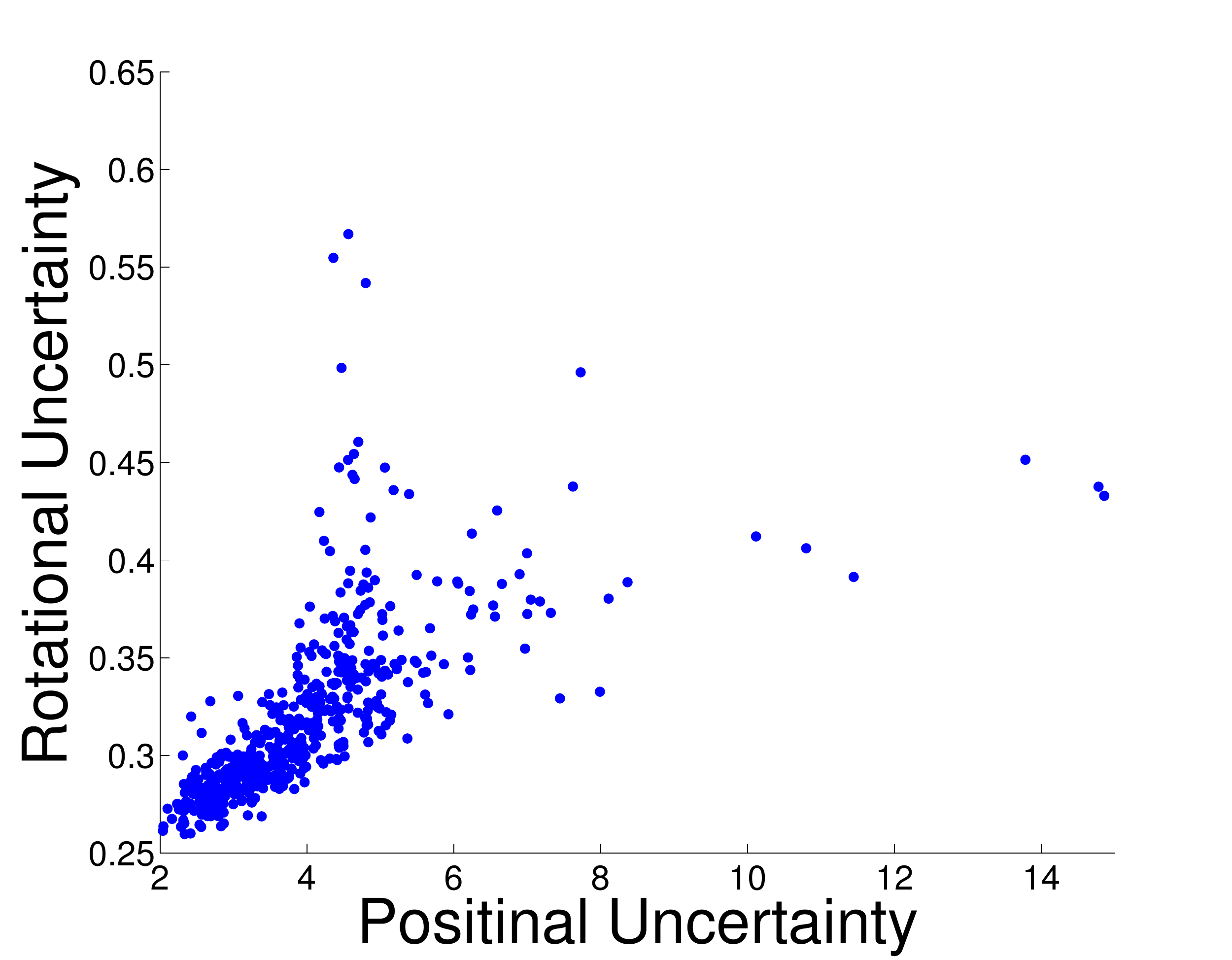}
        \caption{St Mary's Church}
    \end{subfigure}
	\begin{subfigure}[b]{0.33\linewidth}
        \includegraphics[width=\linewidth]{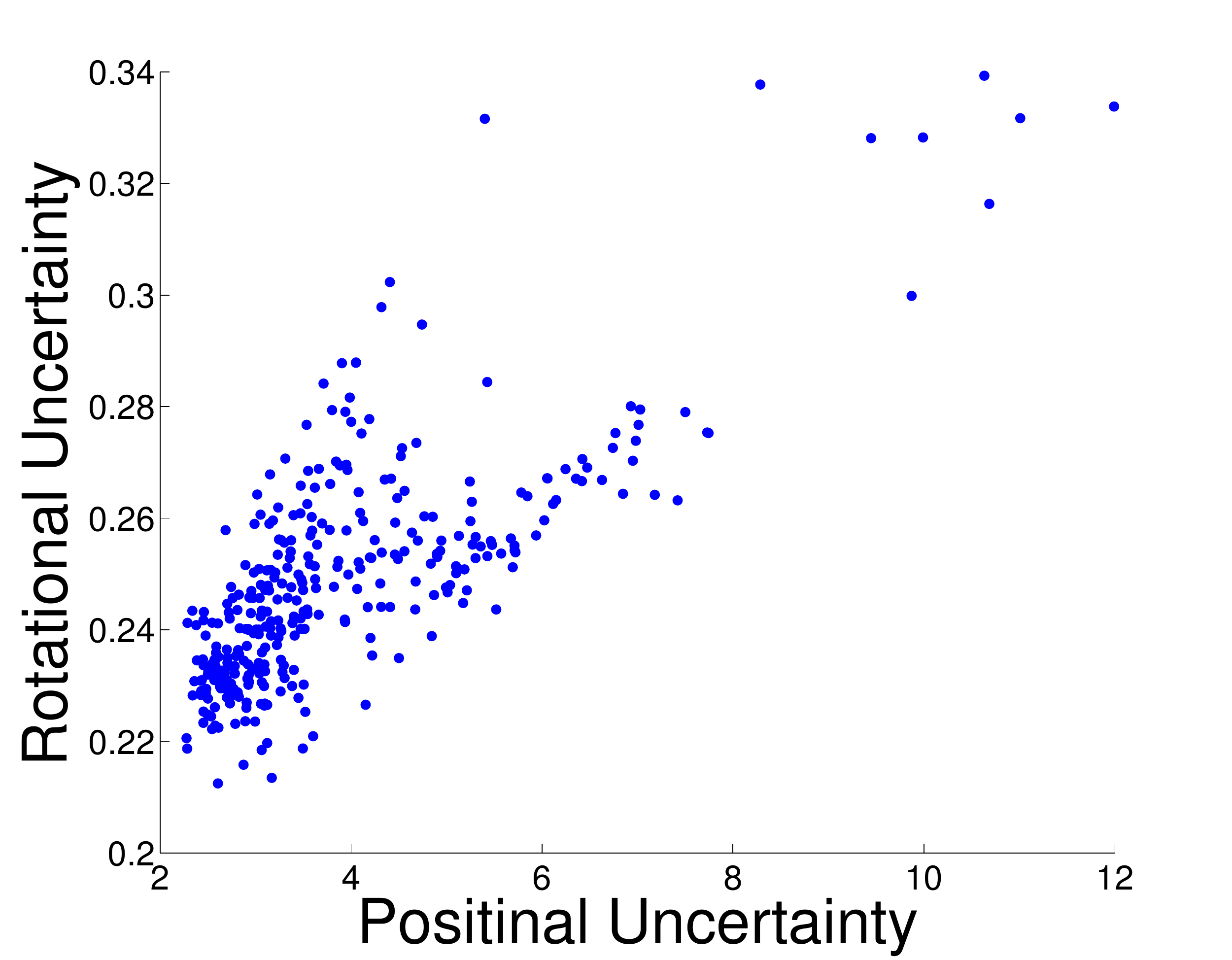}
        \caption{King's College}
    \end{subfigure}
    	\begin{subfigure}[b]{0.33\linewidth}
        \includegraphics[width=\linewidth]{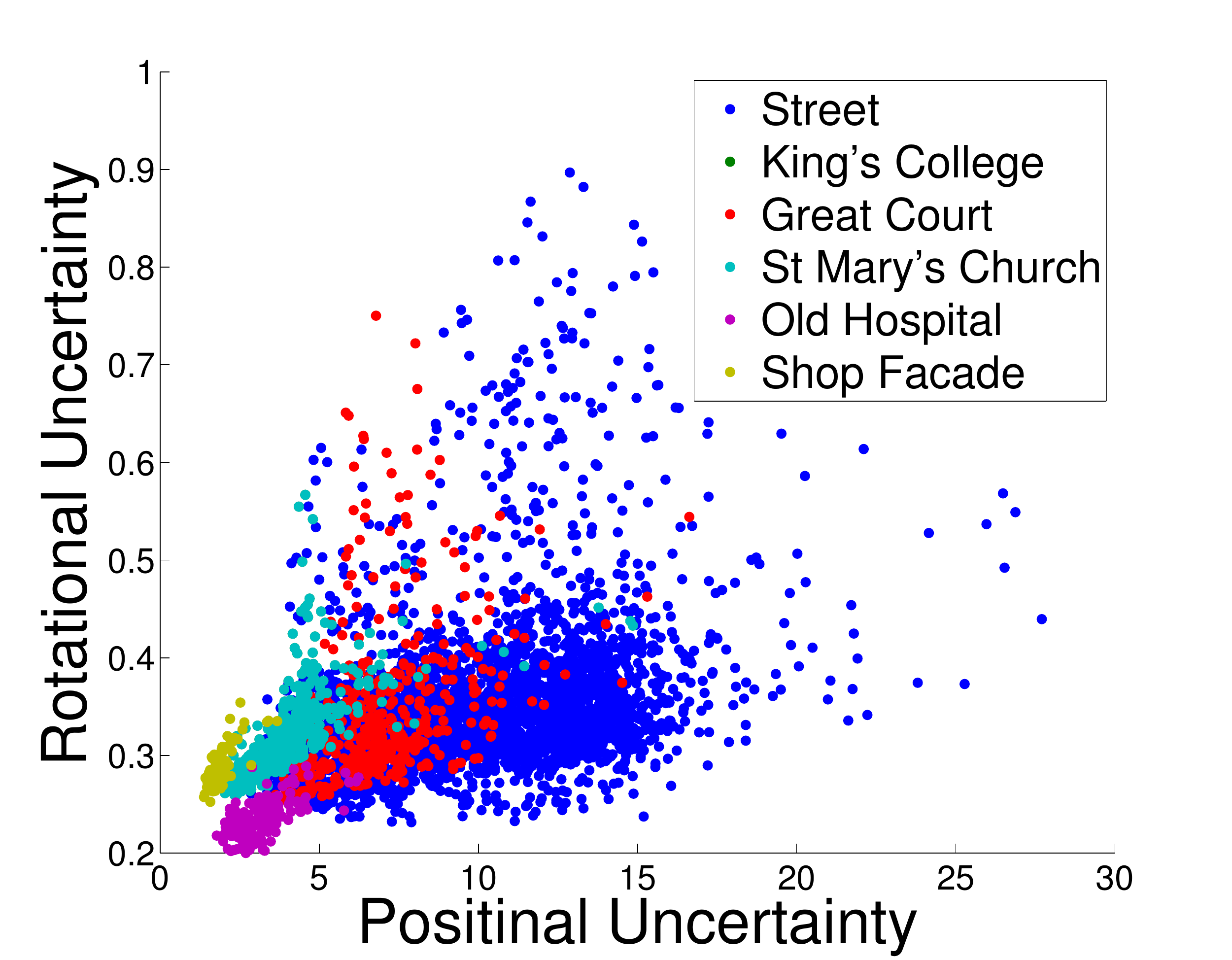}
        \caption{All Scenes}
    \end{subfigure}
    }
\end{center}
   \caption{\textbf{Plot of translational uncertainty against rotational uncertainty} for test images in the St Mary's Church and King's College scene and for all scenes. This shows that the model uncertainty values are very strongly correlated for both rotation and translation. This suggests that we can form a single uncertainty value which represents the overall model uncertainty.}
\label{fig:unc_v_unc}
\end{figure}

Figure \ref{fig:unc_v_unc} shows that the rotational and translational uncertainties are highly correlated. We can therefore compute an overall localization uncertainty by averaging the Z-score for translational and rotational uncertainty. This gives us a final single percentile score which we assign as the confidence of the pose prediction for a given model.

\subsection{Architecture}
\label{ch:arch}

To obtain a fully Bayesian model we should perform dropout sampling after every convolutional layer. However we found in practice this was not empirically optimal. In \cite{kendall2015convolutional} we discovered that fine tuning from pretrained filters trained on a large scale dataset such as \textit{Places} \cite{zhou2014learning} enhanced localization accuracy significantly. This is again true with the probabilistic network. However these pretrained filters were trained without the use of dropout.

Finetuning from weights pretrained on the \textit{Places} \cite{zhou2014learning} dataset, we experimented with adding dropout throughout the model at different depths.  We observe a performance drop in localization when using the fully Bayesian convolutional neural network. Using dropout after every convolutional layer throughout the network acted as too strong a regularizer and degraded performance by approximately 10\% . We obtained the optimal result when we included it only before convolutions which had randomly initialized weights. Therefore we add dropout after inception (sub-net) layer 9 and after the fully connected layers in the pose regressor.

Yosinski et al. \cite{yosinski2014transferable} argues that transferring weights can cause performance to drop in two situations. Firstly when the representation is too specific. However this is unlikely to be the case as we found the weights could successfully generalize to the new task \cite{kendall2015convolutional}. The second explanation was that features may co-adapt fragilely and that transferring them breaks these co-adaptions. We believe this may be the case. The local minima that the weights were optimised to without dropout requires complex co-adaptions that are not able to optimise to a network with the same performance when using dropout.

We did not experiment with changing the dropout probability, or attempt to optimise this hyperparameter. We leave this to future work.

\begin{figure}[t]
\begin{center}
\makebox[\linewidth][c]{
	\begin{subfigure}[b]{0.48\linewidth}
        \includegraphics[width=\linewidth]{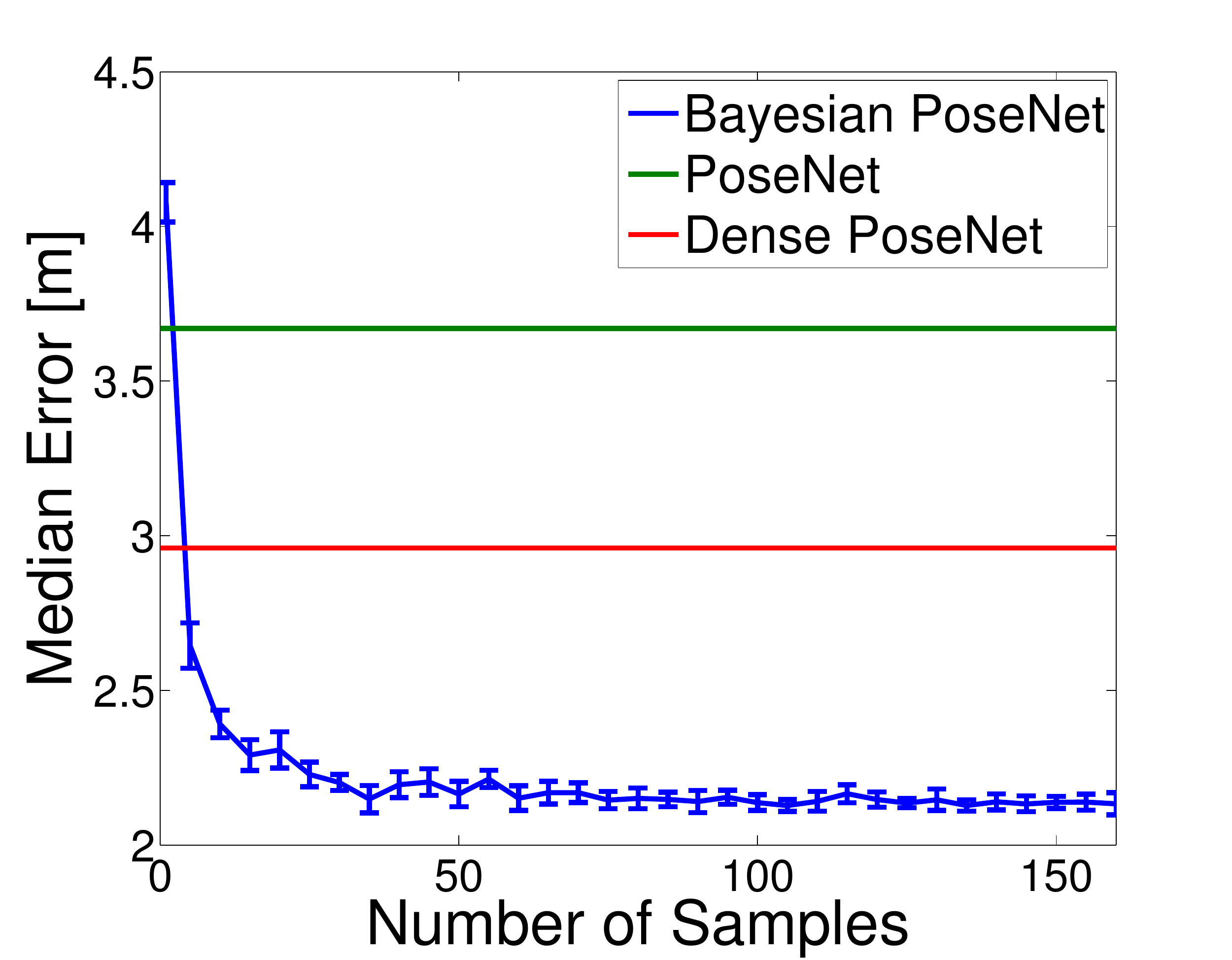}
        \caption{Translation}
    \end{subfigure}
    	\begin{subfigure}[b]{0.48\linewidth}
        \includegraphics[width=\linewidth]{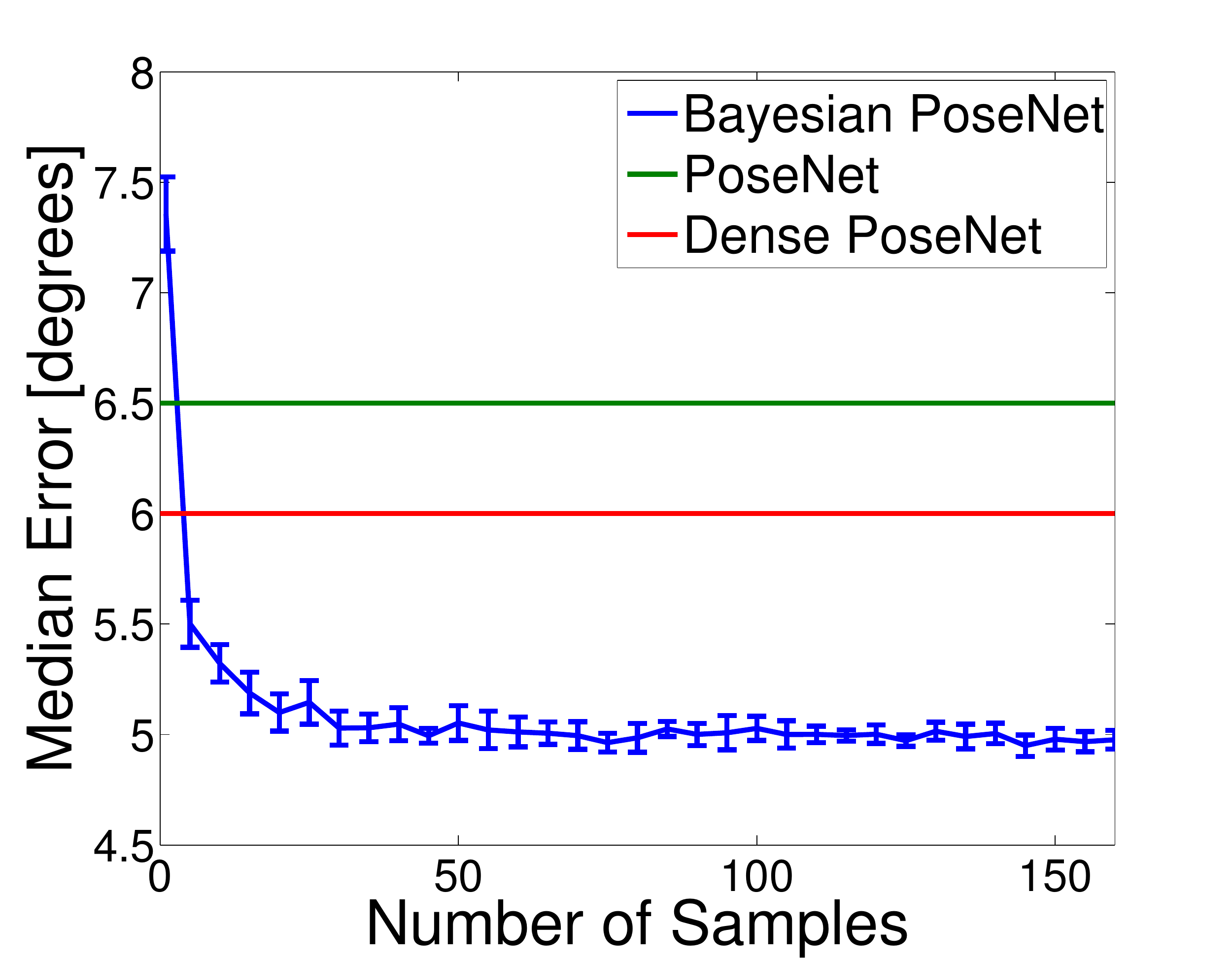}
        \caption{Rotation}
    \end{subfigure}
    }
\end{center}
   \caption{\textbf{Localization accuracy in the Street scene for different number of Monte Carlo samples.} Results are averaged over 8 repetitions, with 1 standard deviation error bars shown. Horizontal lines are shown representing the performance of PoseNet (green) and densely evaluated PoseNet (red) \cite{kendall2015convolutional}. This shows that Monte Carlo sampling provides significant improvement over both these point estimate models after a couple of samples. Monte Carlo sampling converges after around 40 samples and no more significant improvement is observed with more samples.}
\label{fig:samples}
\end{figure}

With this architecture we can then sample from the probabilistic model at test time to obtain an estimate of pose. We can improve localization performance by averaging the Monte Carlo dropout samples \cite{Gal2015Bayesian}. Figure \ref{fig:samples} gives empirical results suggesting that 40 samples are enough to achieve convergence of Monte Carlo samples. We show that less than five samples are typically required to surpass the performance of using a single pose regression convolutional net. After approximately 40 samples no more increase in localization accuracy is observed.

\section{Results}

\begin{figure*}[t]
\begin{center}
\tabcolsep=0.11cm
\begin{tabular}{l|c|c c c c c}
 & Spatial & SCORE Forest & Dist. to Conv. & & & Bayesian \\
Scene & Extent & (Uses RGB-D) & Nearest Neighbour & PoseNet & Dense PoseNet & PoseNet \\
\hhline{=|=|=====}
King's College 		& 140 $\times$ 40m 	& N/A & 3.34m, 5.92\degree & 1.92m, 5.40\degree & 1.66m, 4.86\degree & 1.74m, 4.06\degree\\
Street 				& 500 $\times$ 100m 	& N/A & 1.95m, 9.02\degree & 3.67m, 6.50\degree & 2.96m, 6.00\degree & 2.14m, 4.96\degree\\
Old Hospital 		& 50 $\times$ 40m 	& N/A & 5.38m, 9.02\degree & 2.31m, 5.38\degree & 2.62m, 4.90\degree & 2.57m, 5.14\degree\\
Shop Fa\c cade 		& 35 $\times$ 25m 	& N/A & 2.10m, 10.4\degree & 1.46m, 8.08\degree & 1.41m, 7.18\degree & 1.25m, 7.54\degree\\
St Mary's Church 	& 80 $\times$ 60m 	& N/A & 4.48m, 11.3\degree & 2.65m, 8.48\degree & 2.45m, 7.96\degree & 2.11m, 8.38\degree\\
\hline
Average &  								& N/A & 3.45m, 9.13\degree & 2.40m, 6.76\degree & 2.22m, 6.18\degree & 1.96m, 6.02\degree\\
\hline
\multicolumn{7}{c}{}\\
\hline
Chess 		& 3$\times$2$\times$1m 		& 0.03m, 0.66\degree & 0.41m, 11.2\degree & 0.32m, 8.12\degree & 0.32m, 6.60\degree & 0.37m, 7.24\degree\\
Fire 		& 2.5$\times$1$\times$ 1m 	& 0.05m, 1.50\degree & 0.54m, 15.5\degree & 0.47m, 14.4\degree & 0.47m, 14.0\degree & 0.43m, 13.7\degree\\
Heads 		& 2$\times$0.5$\times$1m 	& 0.06m, 5.50\degree & 0.28m, 14.0\degree & 0.29m, 12.0\degree & 0.30m, 12.2\degree & 0.31m, 12.0\degree\\
Office 		& 2.5$\times$2$\times$1.5m 	& 0.04m, 0.78\degree & 0.49m, 12.0\degree & 0.48m, 7.68\degree & 0.48m, 7.24\degree & 0.48m, 8.04\degree\\
Pumpkin 		& 2.5$\times$2$\times$1m 	& 0.04m, 0.68\degree & 0.58m, 12.1\degree & 0.47m, 8.42\degree & 0.49m, 8.12\degree & 0.61m, 7.08\degree\\
Red Kitchen 	& 4$\times$3$\times$1.5m 	& 0.04m, 0.76\degree & 0.58m, 11.3\degree & 0.59m, 8.64\degree & 0.58m, 8.34\degree & 0.58m, 7.54\degree\\
Stairs 		& 2.5$\times$2$\times$1.5m 	& 0.32m, 1.32\degree & 0.56m, 15.4\degree & 0.47m, 13.8\degree & 0.48m, 13.1\degree & 0.48m, 13.1\degree\\
\hline
Average & 								& 0.08m, 1.60\degree & 0.49m, 13.1\degree & 0.44m, 10.4\degree & 0.45m, 9.94\degree & 0.47m, 9.81\degree\\
\hline
\end{tabular}
\end{center}
\caption{\textbf{Median localization results for the \textit{Cambridge Landmarks} \cite{kendall2015convolutional} and \textit{7 Scenes} \cite{shotton2013scene} datasets.} We compare the performance of the probabilistic PoseNet to PoseNet and a nearest neighbour baseline \cite{kendall2015convolutional}. Additionally we compare to SCORE Forests \cite{shotton2013scene} which requires depth input, limiting it to indoor scenes. The performance of the uncertainty model is shown with 100 Monte Carlo dropout samples. In addition to the qualitative improvement of obtaining an uncertainty metric, we also observe an improvement in relocalization accuracy of up to 10\% over Dense PoseNet.}
\label{tbl:unc_results}
\end{figure*}

We evaluate the performance of the Bayesian convolutional neural network pose regressor on the localization dataset, \textit{Cambridge Landmarks}, which was introduced in \cite{kendall2015convolutional}. Additionally we present results on an indoor relocalization dataset, \textit{7 Scenes} \cite{shotton2013scene}. Table \ref{tbl:unc_results} presents the experimental results of localization accuracy, averaging 100 Monte Carlo dropout samples from the probabilistic PoseNet. We compare this to PoseNet introduced in \cite{kendall2015convolutional} and to a nearest neighbour baseline \cite{kendall2015convolutional} which finds the nearest pose from the training set in feature vector space. We also compare to the SCORE Forest algorithm \cite{shotton2013scene} which is state-of-the-art for relocalization with depth data, however the need for RGB-D data often constrains it to indoor use. 

The results in Table \ref{tbl:unc_results} show that using Monte Carlo dropout \cite{Gal2015Bayesian} results in a considerable improvement in localization accuracy over PoseNet \cite{kendall2015convolutional}. Allowing the model to take into account the uncertainty of model selection, by placing a Bernoulli distribution over the weights, results in more accurate localization. The Monte Carlo samples allow us to obtain estimates of poses probabilistically over the distribution of models. The mean of these samples results in a more accurate solution.

\begin{figure}[t]
\begin{center}
\makebox[\linewidth][c]{
	\begin{subfigure}{0.49\linewidth}
        \includegraphics[width=\linewidth]{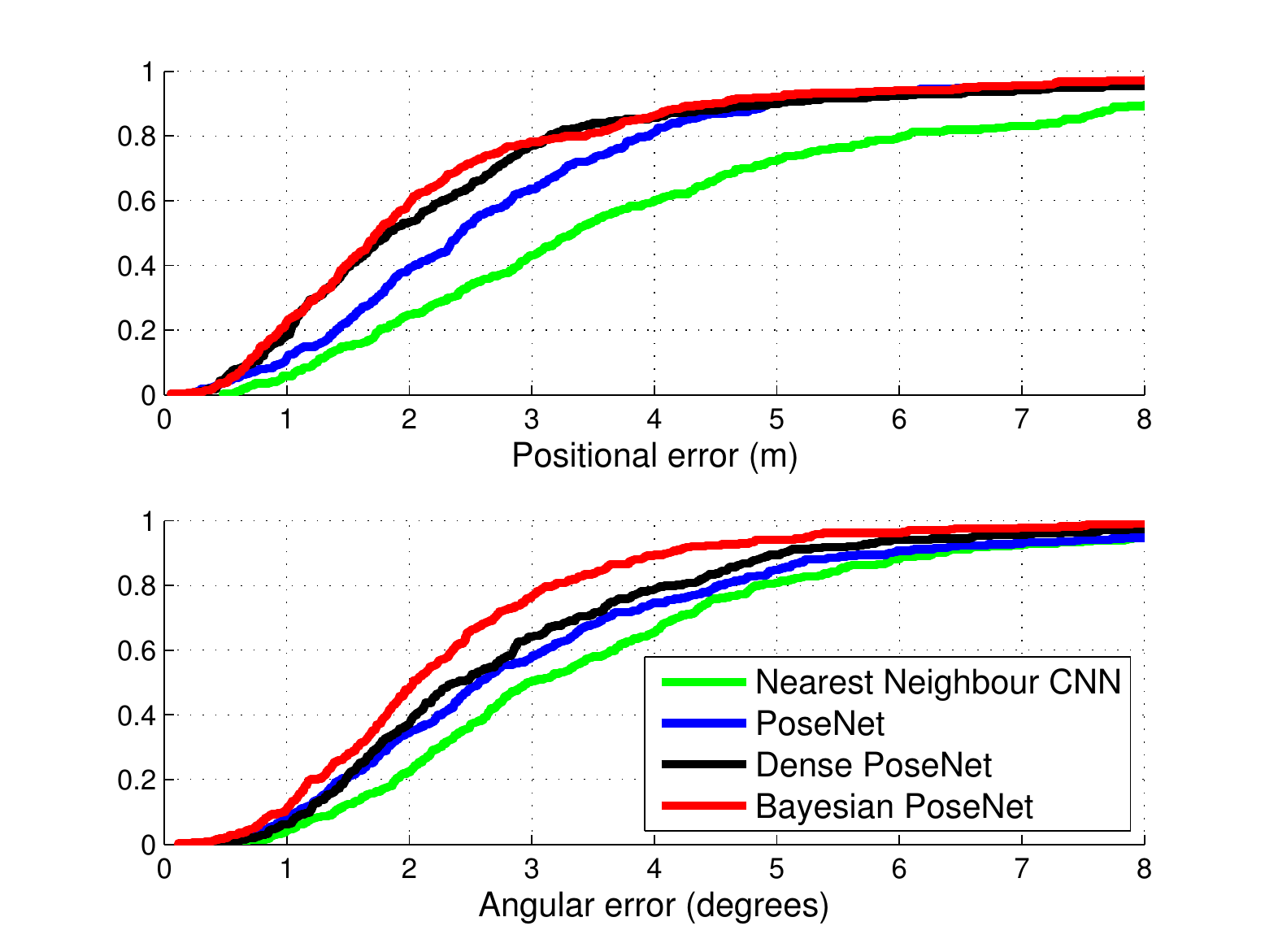}
        \caption{King's College}
    \end{subfigure}
    	\begin{subfigure}{0.49\linewidth}
        \includegraphics[width=\linewidth]{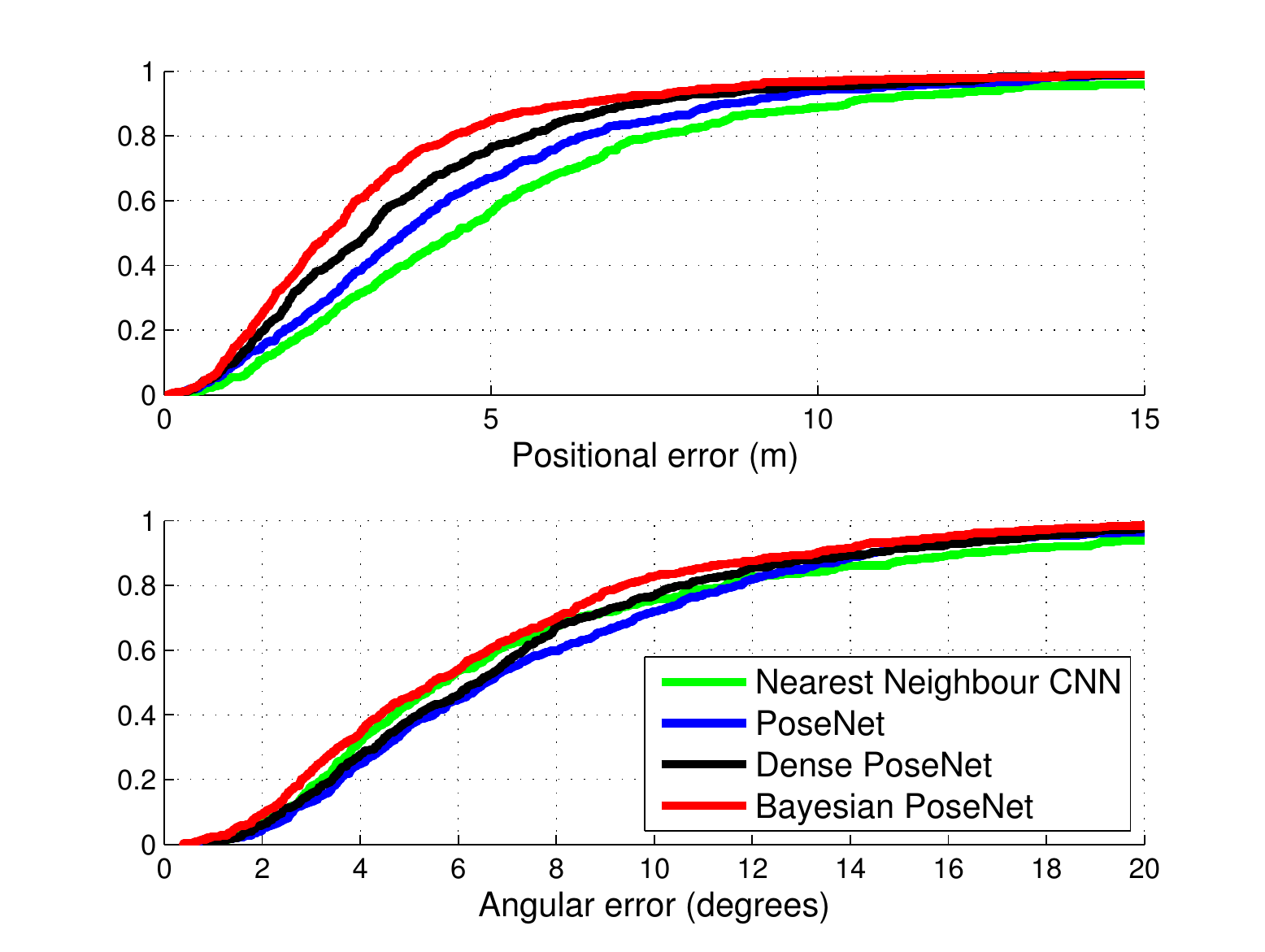}
        \caption{St Mary's Church}
    \end{subfigure}
    }
\end{center}
   \caption{\textbf{Localization accuracy for both position and orientation as a cumulative histogram of errors for the entire test set.} This shows that our probabilistic PoseNet performs consistently better than the non-probabilistic PoseNet for all error thresholds.}
\label{fig:hist}
\end{figure}

Figure \ref{fig:hist} shows a cumulative histogram of errors for two scenes. This shows that our probabilistic PoseNet performs consistently better than the non-probabilistic PoseNet for all error thresholds.

\subsection{Uncertainty as an Estimate of Error}

\begin{figure}[t]
\begin{center}
\makebox[\linewidth][c]{
	\begin{subfigure}[b]{\linewidth}
	\makebox[\linewidth][c]{
        \includegraphics[width=0.5\linewidth]{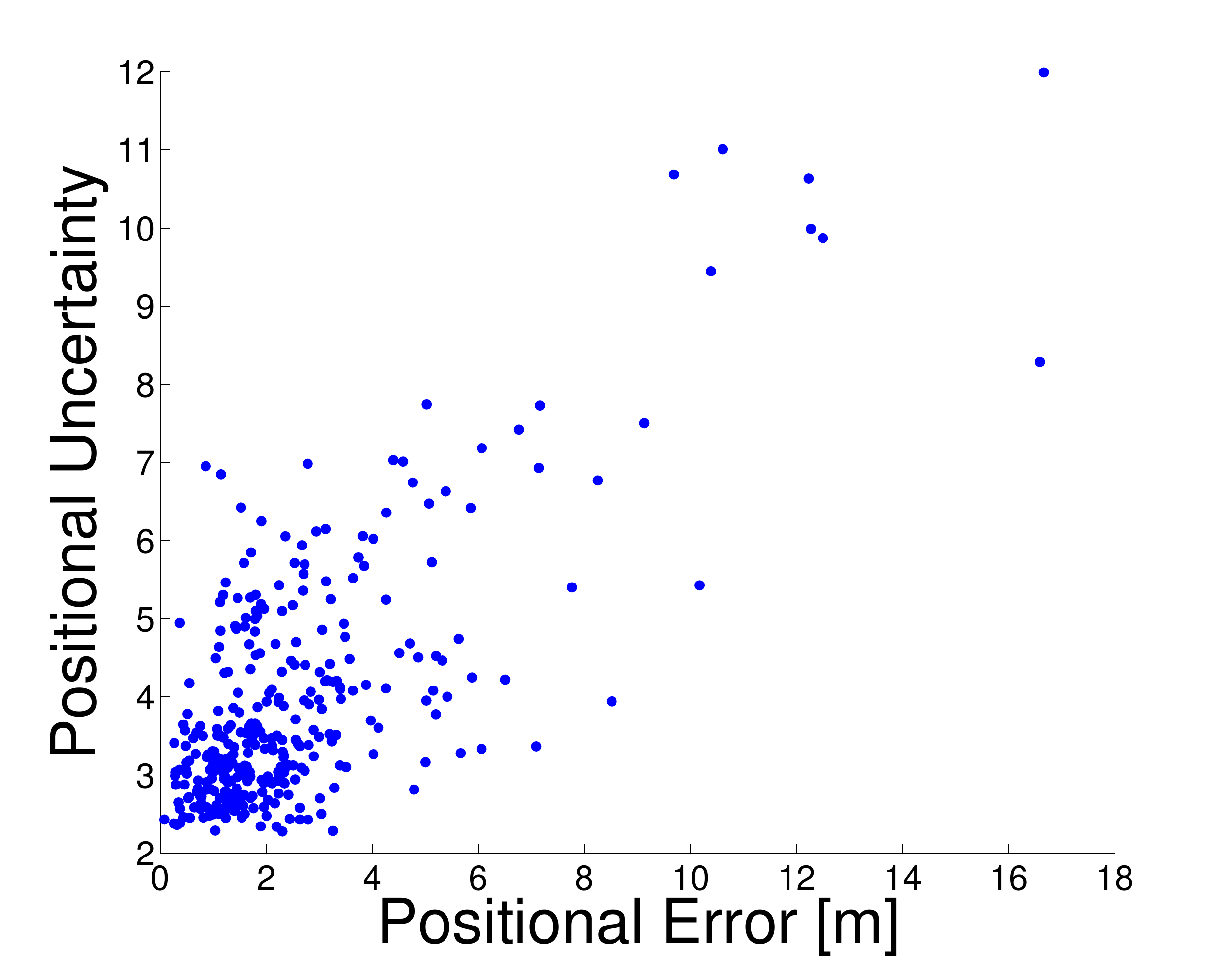}
        \includegraphics[width=0.5\linewidth]{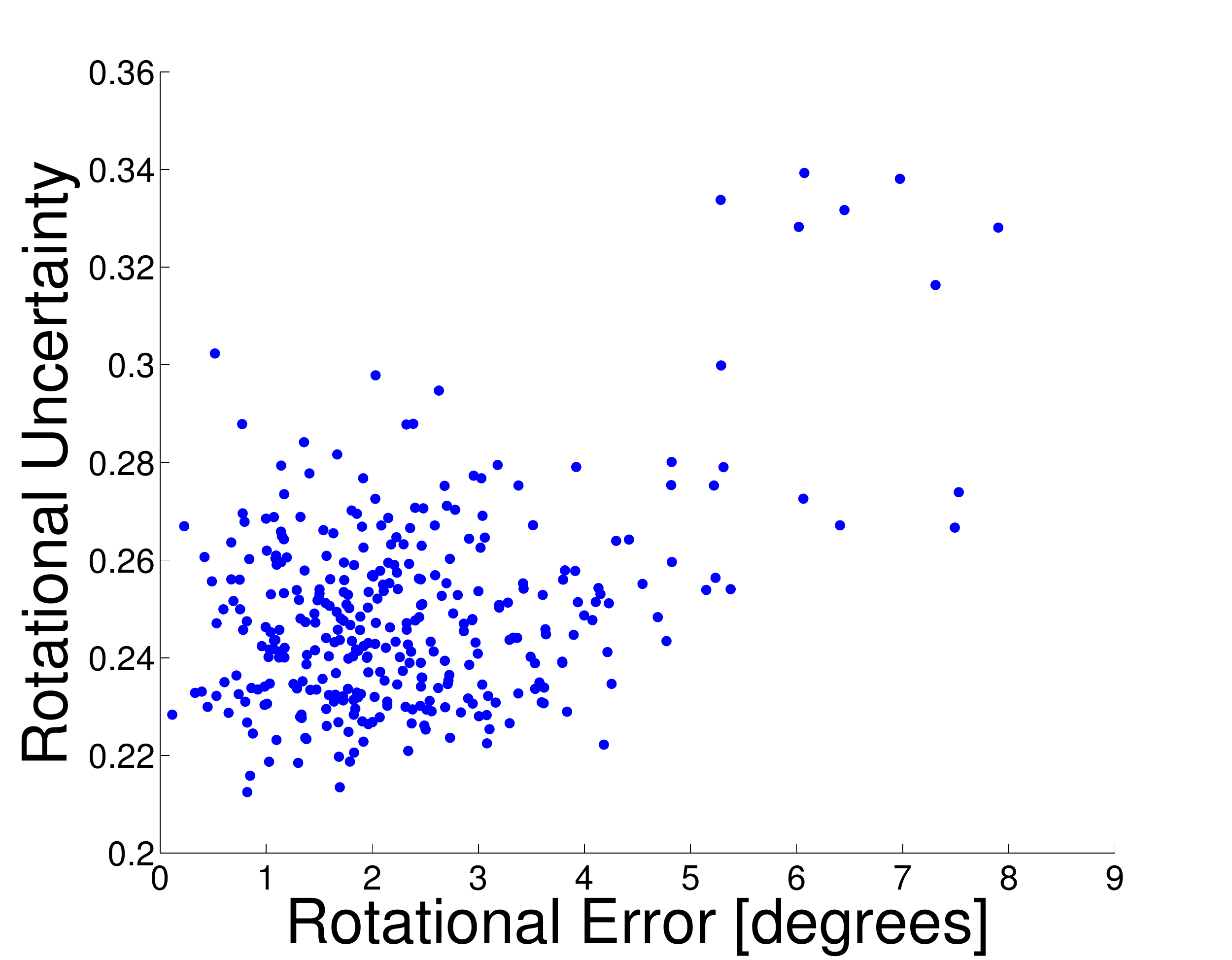}
        }
        \caption{King's College}
    \end{subfigure}
    }
    
\makebox[\linewidth][c]{
	\begin{subfigure}[b]{\linewidth}
	\makebox[\linewidth][c]{
        \includegraphics[width=0.5\linewidth]{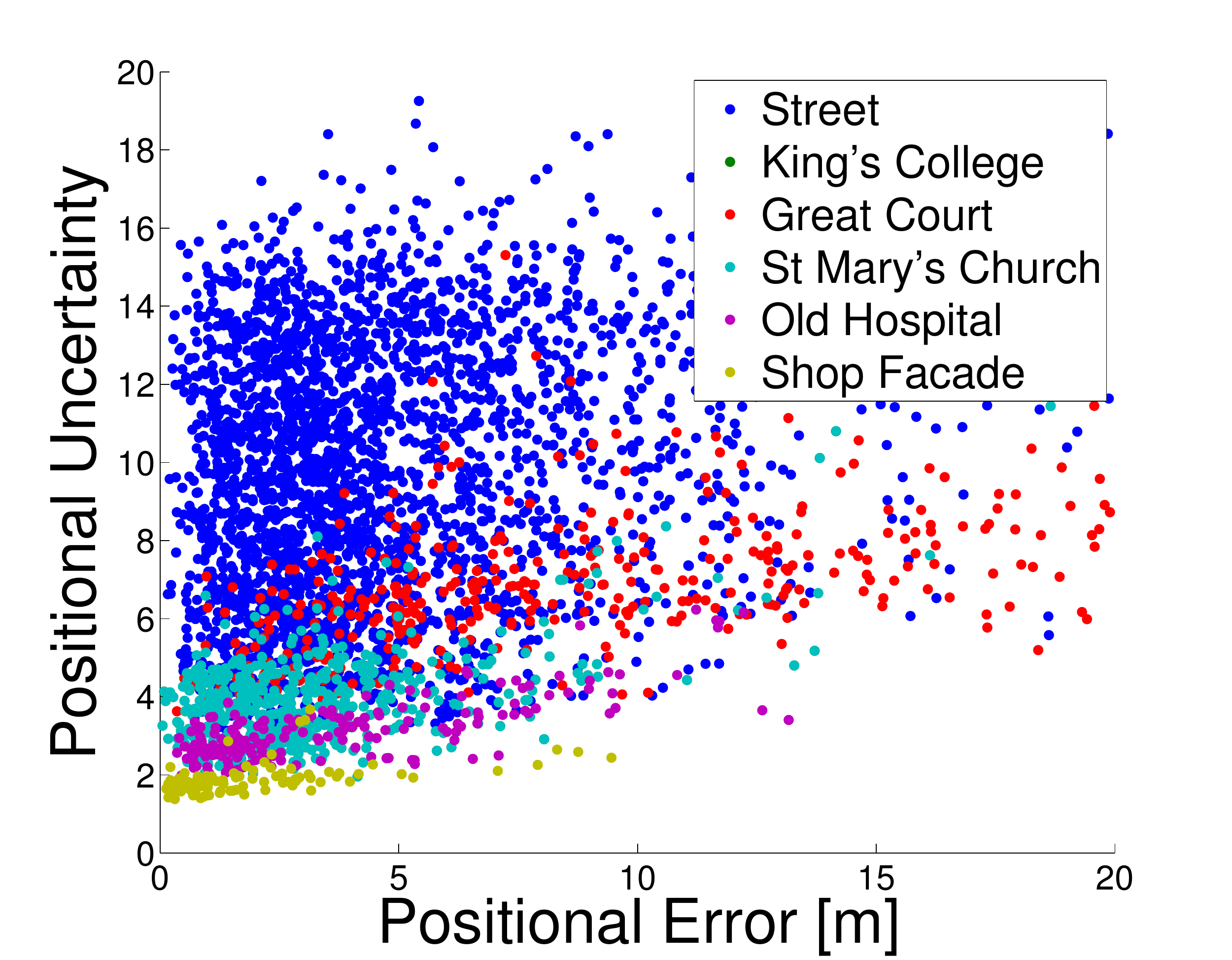}
        \includegraphics[width=0.5\linewidth]{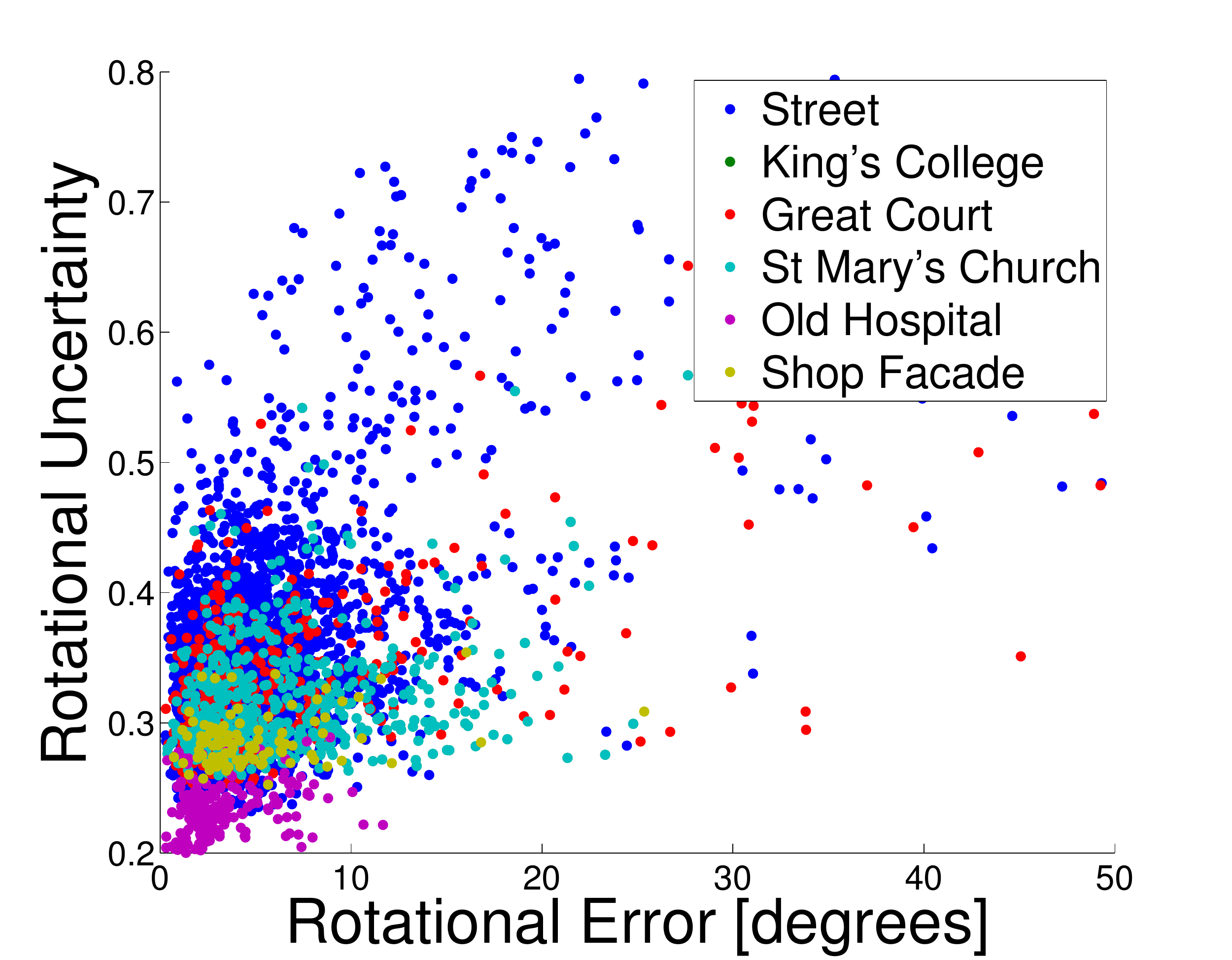}
        }
        \caption{All Scenes}
    \end{subfigure}
    }
\end{center}
   \caption{\textbf{Plot of translational and rotational errors against their respective estimated uncertainty} for test images in the King's College scene and for all scenes. These plots show that the uncertainty is correlated with error and provides a good estimate of metric relocalization error. It also shows that the scale of uncertainty values that each model learns varies significantly, suggesting they should be normalized for each model, as proposed in Section \ref{ch:uncertainty_dist}.}
\label{fig:error_vs_uncertainty}
\end{figure}

Figure \ref{fig:error_vs_uncertainty} shows that the uncertainty estimate is very strongly correlated with metric relocalization error. This shows that we can use the uncertainty estimate to predict relocalization error. The plot shows that this relationship is linear for both translational and rotational uncertainty. However the proportionality gradient between error and uncertainty varies significantly between scenes.

Figure \ref{fig:unc_v_unc} shows that metric error and uncertainty values are correlated between rotational and translational values. This supports the assumptions in our method of generating an overall uncertainty estimate as an `average' of these normalized values. We observe relocalization error and uncertainty are strongly correlated between both position and orientation.

\subsection{Uncertainty as a Landmark Detector}

\begin{figure}[t]
\begin{center}
\begin{subfigure}[b]{\linewidth}
   	\includegraphics[width=\linewidth]{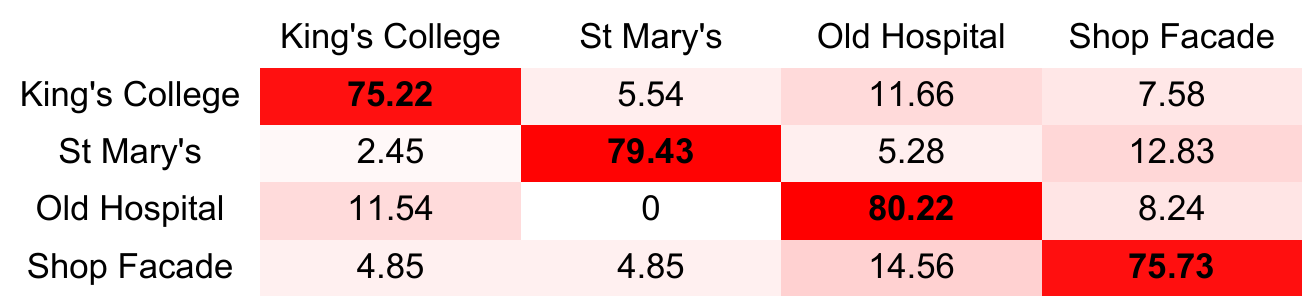}
   \caption{Confusion matrix for \textit{Cambridge Landmarks} dataset}
\end{subfigure}
\begin{subfigure}[b]{\linewidth}
   	\includegraphics[width=\linewidth]{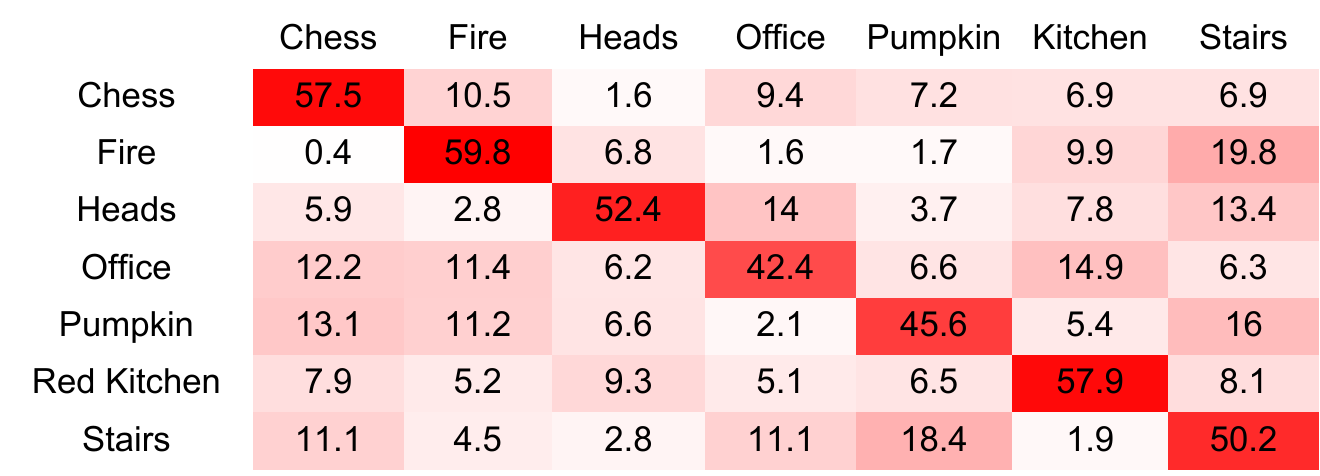}
   \caption{Confusion matrix for \textit{7 Scenes} dataset}
\end{subfigure}
\end{center}
   \caption{\textbf{Scene recognition confusion matrices.} For each dataset (row) we computed the Z-score for both rotation and translation uncertainties. Dataset images were classified to the model (column) with the lowest uncertainty. Note that the Street scene is excluded as it contains many of the other landmarks in \textit{Cambridge Landmarks}. This shows that the uncertainty metric is able to recognize correctly the landmark that it was trained to relocalize from. The network outputs large model uncertainty when it is presented with an unknown scene. The average scene detection accuracy is approximately 78\% for \textit{Cambridge Landmarks}. The indoor dataset is a far more challenging problem, as many scenes are very visually ambiguous. For example the pumpkin scene is the same room as the kitchen, with a different arrangement. Despite this, our system still performs modestly with 52\% accuracy.}
\label{fig:confusion_matrix}
\end{figure}

We show that the uncertainty metric can also be used to determine if the image is from the scene or landmark that the pose regressor was trained on. For a given scene in a dataset we test each image on all of the models. We then compute the uncertainty metric using the normalization method proposed in Section \ref{ch:uncertainty_dist}. The image should have the lowest uncertainty value with the model which was trained on the scene that the image was taken from.

In Figure \ref{fig:confusion_matrix} we present a confusion matrix showing this result for the \textit{Cambridge Landmarks} and \textit{7 Scenes} datasets. We exclude the Street scene as it contains many of the landmarks in the other scenes. We show the confusion matrix when using the combined normalized uncertainty. We observed that combining the rotation and translation metrics often provides a superior and more robust error metric.

Note that the network has not been trained to classify the landmark it is observing. This is obtained as a by-product of the probabilistic architecture. If the convolutional net was trained to classify landmarks we are confident that it would perform significantly better. The purpose of this was to validate that the uncertainty measurement can reflect whether or not the network is trained on what it is presented with. The results show that the system can not only estimate the accuracy of the prediction, but also correctly identify when the landmark is not present at all.

\subsection{What Makes the Model Uncertain About a Pose?}

An initial hypothesis may be that test images which are far from training examples give very uncertain results, because they are more unknown to the network. To study this we plotted, for each test image in a scene, the uncertainty against the Euclidean distance between the test image and the nearest training image. This plot showed a very slight increasing trend but was not sufficiently clear to draw any conclusions. However Euclidean distance to the nearest training image is not a comprehensive measure of similarity to the training set. It excludes other variables such as orientation, weather, pedestrian activity and lighting. 

PoseNet produces a 2048 dimensional feature vector (see Section 3.2 of \cite{kendall2015convolutional}). This feature vector contains a high dimensional representation of instantiation parameters in the scene, such as weather, lighting, and object pose. Therefore we use this feature vector as a representation of the image. To compute similarity between two images, we evaluate the pose regressor's feature vector for each image and take the Euclidean distance between each feature vector. Therefore we can use this as a measure of similarity between a test image and the dataset's training image by finding the distance to the nearest neighbour training image in this feature space. This is the same measure used to compute the nearest neighbour results in Table \ref{tbl:unc_results}.

\begin{figure}[t]
\begin{center}
   	\includegraphics[width=0.75\linewidth]{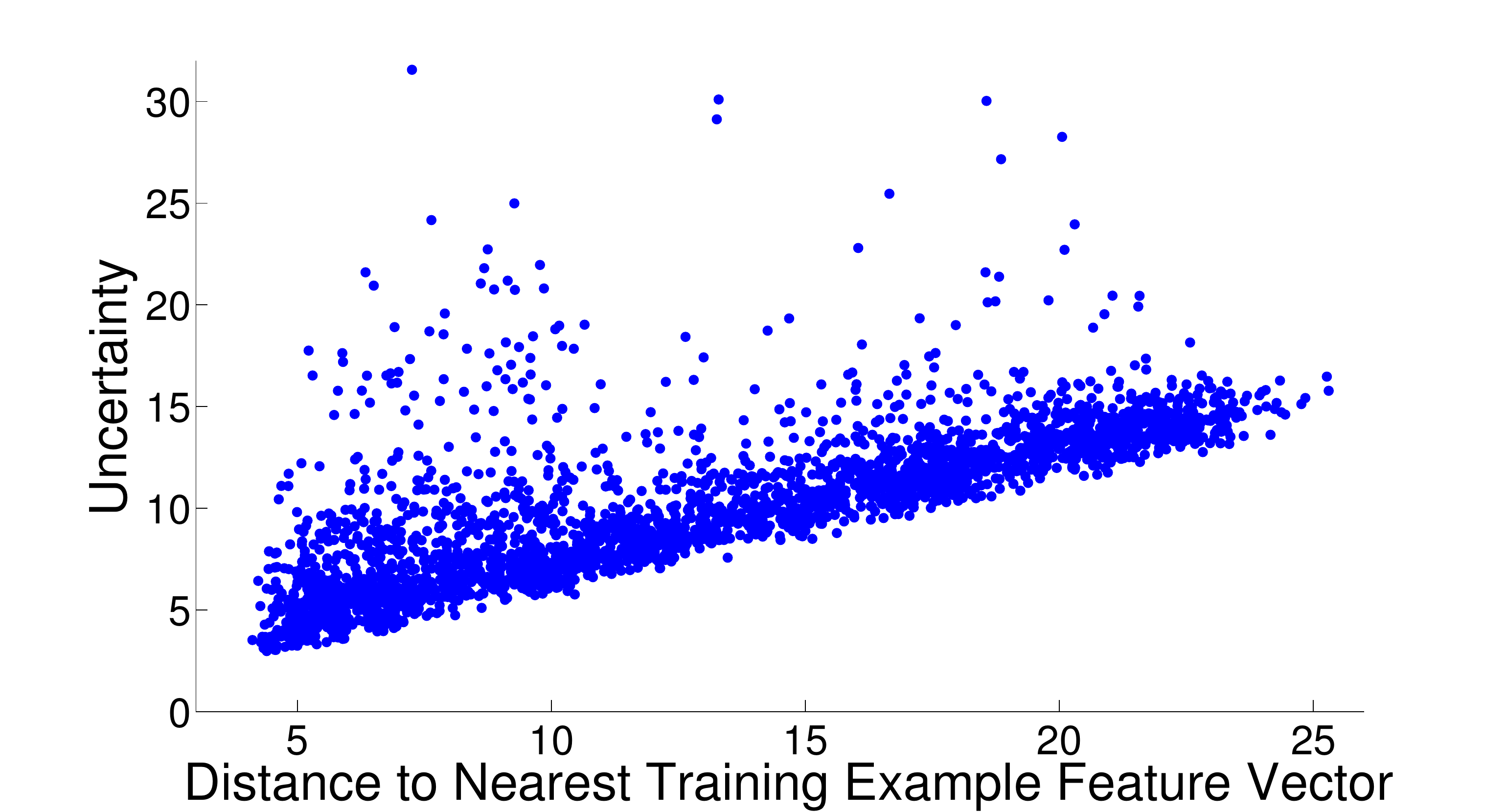}
\end{center}
   \caption{\textbf{Uncertainty value for test images in the Street scene, plotted against Euclidean distance to the nearest neighbour training image feature vector.} The feature vector is a 2048 dimensional vector obtained from the final layer in PoseNet before the pose regression. This shows that having similar training examples lowers model uncertainty in test images.}
\label{fig:nn_vec_uncertainty}
\end{figure}

Figure \ref{fig:nn_vec_uncertainty} shows a plot of model uncertainty against this distance for all test images in the Street scene. The strong relationship indicates that the model is more uncertain for images which are less similar (in this localization feature space) to those in the training dataset. 

\begin{figure*}[t]
\begin{center}
\makebox[\linewidth][c]{
   	\includegraphics[width=0.14\linewidth]{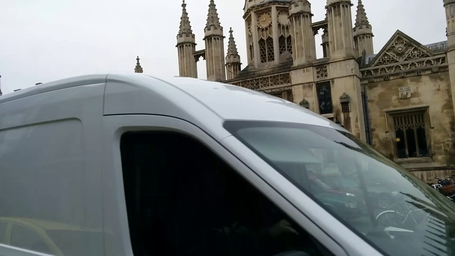}
   	\includegraphics[width=0.14\linewidth]{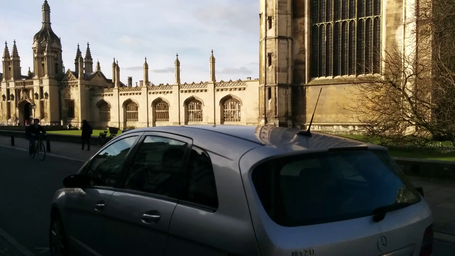}
   	\includegraphics[width=0.14\linewidth]{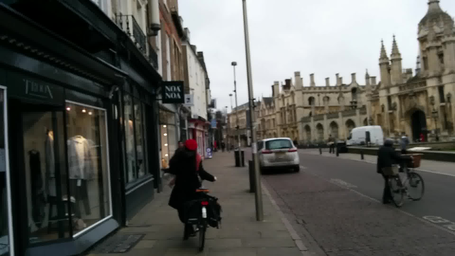}
   	\includegraphics[width=0.14\linewidth]{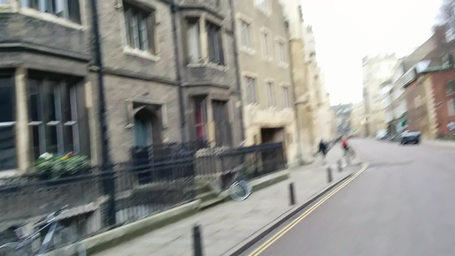}
   	\includegraphics[width=0.14\linewidth]{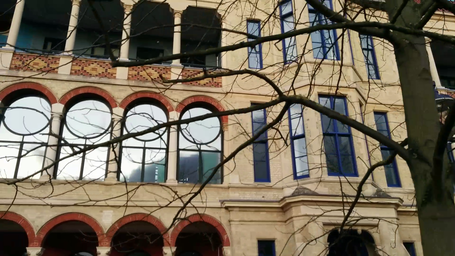}
   	\includegraphics[width=0.14\linewidth]{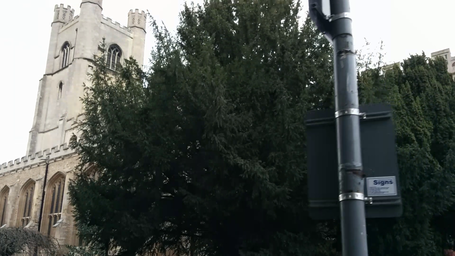}
   	}
   	
   	\vspace{1mm}
   	
\makebox[\linewidth][c]{
   	\includegraphics[width=0.14\linewidth]{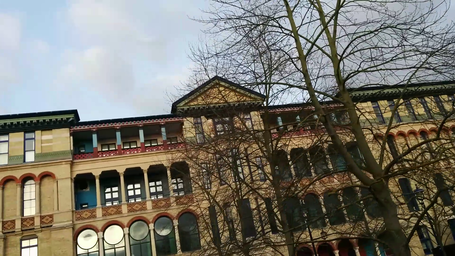}
   	\includegraphics[width=0.14\linewidth]{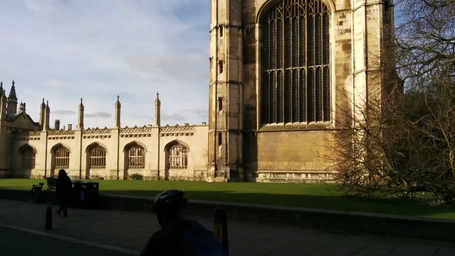}
   	\includegraphics[width=0.14\linewidth]{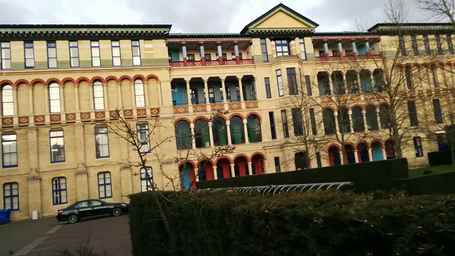}
   	\includegraphics[width=0.14\linewidth]{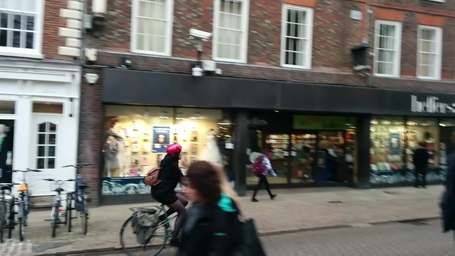}
   	\includegraphics[width=0.14\linewidth]{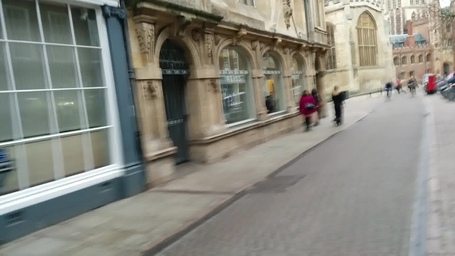}
   	\includegraphics[width=0.14\linewidth]{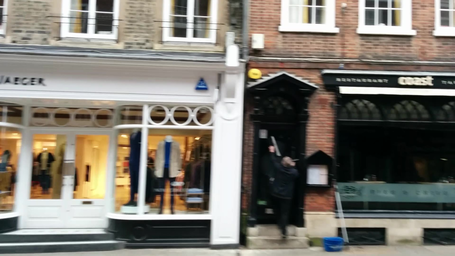}
   	}
\end{center}
   \caption{\textbf{Images with the largest uncertainty values and largest localization errors.} All of these images contain one of the following situations causing difficult and uncertain localization: strong occlusion from vehicles, pedestrians or other objects, motion blur, are taken from an area at the edge of the scene or are distant from a training example.}
\label{fig:difficult_examples}
\end{figure*}

The points which deviate from this trend, with larger uncertainty values, are typically the difficult images to localize. Some examples are shown in Figure \ref{fig:difficult_examples} These images have challenges such as heavy vehicle occlusion or strong silhouette lighting which result in inaccurate and uncertain prediction.

%

\subsection{System Efficiency}

We now compare the performance of our probabilistic PoseNet to our non-probabilistic PoseNet introduced in \cite{kendall2015convolutional}. The probabilistic approach shares all the same benefits of PoseNet \cite{kendall2015convolutional}, being scalable as its memory and computational requirements do not scale with map size or training data. Introducing dropout uncertainty does not require any more parametrisation and the weight file remains constant at $50$ MB. This is still much more efficient than the gigabytes required for metric relocalization systems with point features \cite{li2012worldwide}.

Drawing stochastic samples comes at a small additional time cost. As Figure \ref{fig:samples} shows, the optimal samples to take is approximately 40 as any more samples than this does not significantly improve performance. When operating on a parallel processor, such as a GPU, this extra computation is manageable by treating it as a mini-batch of operations. Only the final two fully connected layers need to be sampled, as the first 21 convolutional layers are deterministic. For example, on an NVIDIA Titan X, computing pose by averaging 40 Monte Carlo dropout samples takes $5.4ms$ while 128 samples takes $6ms$. For comparison, a single PoseNet evaluation takes $5ms$ per image.

\section{Conclusions}

We show how to successfully apply an uncertainty framework to the convolutional neural network pose regressor, PoseNet. This improved PoseNet's real time relocalization accuracy for indoor and outdoor scenes. We do this by averaging \textit{Monte Carlo dropout} samples from the posterior Bernoulli distribution of the Bayesian convolutional network's weights. This is requires no extra parametrisation.

Furthermore we show the trace of these sample's covariance matrix provides an appropriate model uncertainty estimate. We show that this uncertainty estimate accurately reflects the metric relocalization error and can be used to detect the presence of a previously observed landmark. We present evidence that shows the model is more uncertain about images which are dissimilar to the training examples.

\bibliographystyle{unsrt}
\bibliography{root}

\end{document}